\documentclass[12pt]{article}
\usepackage{amsmath}
\usepackage{amsfonts}
\usepackage{amssymb}
\usepackage{graphicx}
\usepackage{subfig}
\usepackage{setspace}
\usepackage{multirow}
\usepackage{url}
\usepackage[table,xcdraw]{xcolor}

\newtheorem{theorem}{Theorem}[section]
\newtheorem{lemma}{Lemma}[section]
\newtheorem{corollary}{Corollary}[section]

\newtheorem{example}{Example}[section]

\newtheorem{remark}{Remark}[section]

\DeclareGraphicsExtensions{.jpg,.pdf,.png,.jpg,.eps}

\numberwithin{equation}{section}
\newcommand{\bxi}{{\boldsymbol \xi}}
\newcommand{\bet}{{\boldsymbol \eta}}

\def\by{{\boldsymbol y}}

\def\bY{{\boldsymbol Y}}

\def\bx{{\boldsymbol x}}
\def\bX{{\boldsymbol X}}

\def\cM{{\mathcal M}}
\def\cL{{\mathcal L}}
\def\cC{{\mathcal C}}
\def\cS{{\mathcal S}}
\def\cF{{\mathcal F}}
\def\cJ{{\mathcal J}}
\def\cG{{\mathcal G}}
\def\cS{{\mathcal S}}
\def\bbR{{\mathbb R}}

\graphicspath{%
    {Fig/} 
    {/} 
}

\begin{document}

\title{Geometry and clustering with metrics \\ derived from separable Bregman divergences} 

\author{Erika Gomes-Gon\c{c}alves$^1$, Henryk Gzyl$^2$ and Frank Nielsen$^3$\\
\noindent 
$^1$Independent Consultant, Madrid, Spain.\\
$^2$Centro de Finanzas IESA, Caracas, Venezuela.\\
$^3$Sony Computer Science Laboratories, Inc, Japan.\\
 {\small {\tt erikapat@gmail.com}, {\tt  henryk.gzyl@iesa.edu.ve}, {\tt Frank.Nielsen@acm.org}} }

\date{}
 \maketitle

\setlength{\textwidth}{4in}
\vskip 1 truecm
\baselineskip=1.5 \baselineskip \setlength{\textwidth}{6in}
%\newpage
\begin{abstract}
Separable Bregman divergences induce Riemannian metric spaces that are isometric to the Euclidean space after monotone embeddings.
We investigate fixed rate quantization and its codebook Voronoi diagrams,
and report on experimental performances of partition-based, hierarchical, and soft clustering algorithms with respect to these Riemann-Bregman distances.
\end{abstract}

\noindent {\bf Keywords}: Bregman divergence, Riemannian distance, Riemann-Bregman distance, Legendre transformation, Geometry, Voronoi cells, Clustering, Quantization.\\
\noindent{MSC 2000 Subject Classification}.

\begin{spacing}{0.01}
   \tableofcontents
\end{spacing}

%%%%
\section{Introduction and Preliminaries}
%%%%%
\sloppy

%%%%%%%%
\subsection{Riemannian geometry induced by a Bregman divergence}
%%%%%%%%%%
In \cite{G} it was shown how to put a Riemannian structure on a convex open subset $\cM$ of $\mathbb{R}^K$  derived from a Bregman  divergence 
$$
\delta_\Phi(\bx,\bx')=\Phi(\bx)-\Phi(\bx')-(\bx-\bx')^\top\nabla\Phi(\bx'),
$$ 
with real-valued generator function $\Phi:\cM \to \bbR$. 
There $\cM=\cJ^K$ where $\cJ$ is an interval (bounded or unbounded) in the real line, 
and the {\em separable} Bregman generator $\Phi(\bx)=\sum_{j=1}^K\phi(x_j)$ with $\phi:\cJ \to \bbR$ is a three times continuously differentiable function. This set up was chosen for computational convenience, and $\delta_\Phi(x,x') = \sum_{j=1}^K \delta_\phi(x_j,x'_j)$. 
In this case the Riemannian metric on $\cM$ is given by $g_{i,j}(\bx)=\phi''(x_i)\delta_{i,j}$. 

The Riemannian distance in $\cM$ induced by that metric is given by
\begin{equation}\label{dist}
d_\phi(\bx,\bx')^2 = \sum_{j=1}^K \big(h(x_j) - h(x_j')\big)^2,
\end{equation} 
\noindent where $h(x)=\int \sqrt{\phi''(x)}$ is a an increasing continuously differentiable function defined as the primitive of $(\phi'')^{1/2}$ (also called antiderivative). Its compositional inverse $h^{-1}$ will be denoted by $H$ (with $H\circ h^{-1}=h^{-1}\circ H=\mathrm{id}$) and we shall use the same letter for the extensions of the functions to $\cM$, that is, we shall write $h(\bx)$ to denote the transpose of $(h(x_1),...,h(x_K)),$ and the same for $H$. We term distance $d_\phi$ the {\em Riemann-Bregman distance}.

The equation $\gamma(t)=(\gamma_1(t),\ldots,\gamma_K(t))$ of the Riemannian geodesic between $\bx$ and $\bx'$ was reported in \cite{G} as
$\gamma_i(t)=h^{-1}(h(x_i)+k_i t)$,
where the $k_i=h(x_i')-h(x_i)$ are the constants of integration such that $\gamma(0)=\bx$ and $\gamma(1)=\bx'$.
That is, the geodesics are expressed as
\begin{equation}\label{geo}
\gamma_i(t)=h^{-1}\Big((1-t)h(x_i)+ t h(x_i')\Big),\quad t\in[0,1].
\end{equation} 

Notice that by introducing the Legendre convex conjugate 
\begin{eqnarray}
\phi^*(y) &=&\sup_{x\in\cJ} \{yx-\phi(x)\},\\
&=& y{\phi'}^{-1}(y)-\phi({\phi'}^{-1}(y)),
\end{eqnarray}
with $y_i=\phi'(x_i)$ (and $x_i={\phi'}^{-1}(y_i)={\phi^*}'(y_i)$), 
we can write the dual Riemannian metric as ${g^*}^{ij}(\mathbf{y})= {\phi^*}''(y_i)\delta_{i,j}$, and
we have the following identity: $g(\mathbf{x}){g^*}(\mathbf{y})=\mathrm{id}$.

It follows that we can express the Bregman-Riemannian distance either by using the primal affine $\mathbf{x}$-coordinate system, or by using the dual affine $\mathbf{y}$-coordinate system (with $y=\nabla\Phi(x)$ and $x=\nabla\Phi^*(y)$):

\begin{lemma}[Dual Riemann-Bregman distances]
We have $\delta_\Phi(\bx,\bx')=\delta_{\Phi^*}(\by',\by)$, and 
\begin{equation}\label{dualmetric}
d_\phi(\bx,\bx') = d_{\phi^*}(\by,\by') = d_{\phi^*}(\nabla\Phi(\bx),\nabla\Phi(\bx')).
\end{equation}
\end{lemma}

%Indeed, the Riemann-Bregman distance can be expressed in a Bregman manifold either by using the primal affine coordinate system $x$, or equivalently by using the dual affine coordinate system $y=\nabla\Phi(x)$.

\begin{example}
Consider $\phi(x)=x\log x-x$ (extended Shannon negentropy). 
We have $\phi'(x)=\log x$ and $\phi''(x)=\frac{1}{x}$.
It follows that $h(x)=2\sqrt{x}+c_x$, where $c_x$ is a constant.
The Legendre conjugate is $\phi^*(y)=e^y$ and $h^*(y)=\int\sqrt{{\phi^*}''}=2e^{\frac{y}{2}}+c_y$, where $c_y$ is a constant.
We have $y(x)=\log x$,  $d_\phi(x,x')=2|\sqrt{x}-\sqrt{x'}|$ and $d_{\phi^*}(y,y')=2|e^{\frac{y}{2}}-e^{\frac{y'}{2}}|$.
We check that $d_\phi(x,x') = d_{\phi^*}(y,y')$ since  $e^{\frac{y}{2}}=\sqrt{x}$.
\end{example}

\begin{figure}
\centering
\includegraphics[width=11cm]{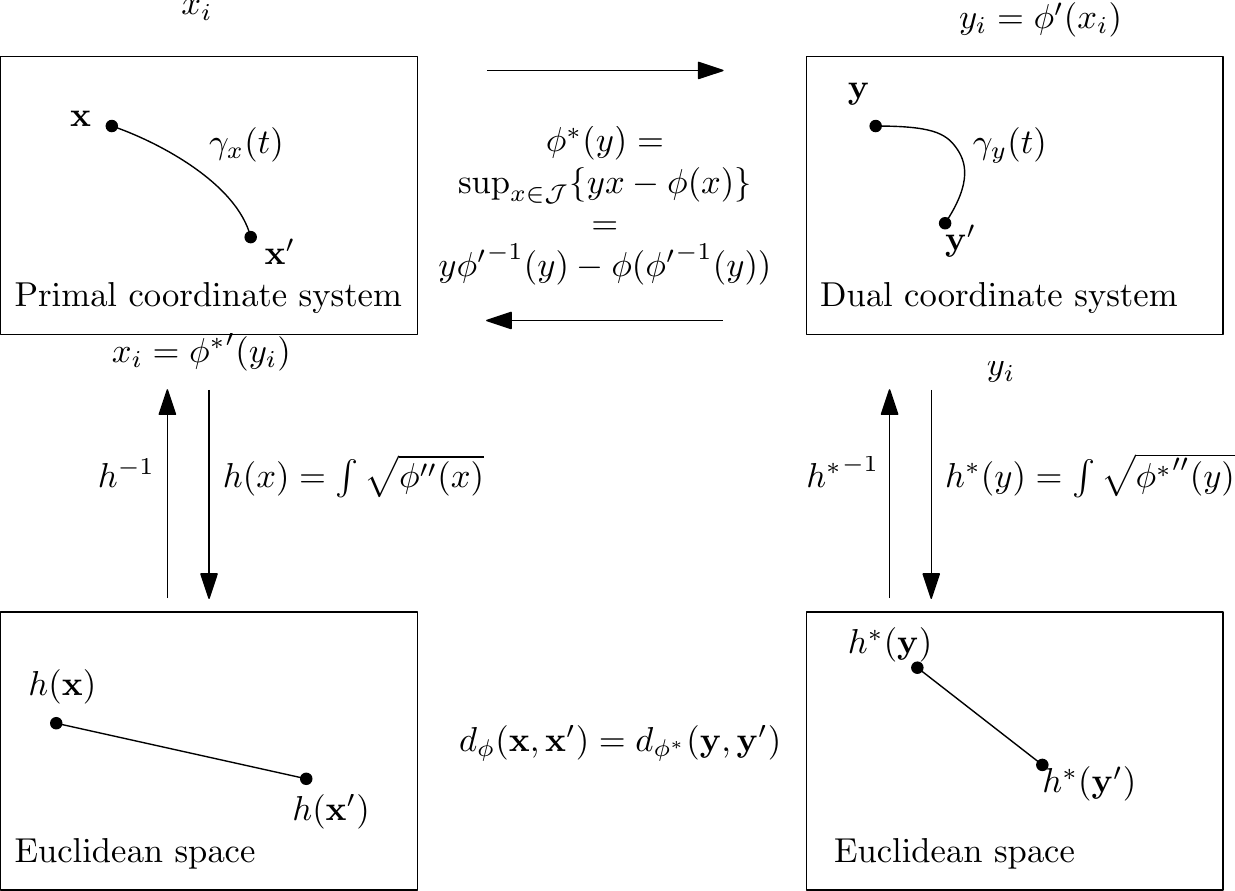}
\caption{The Riemann-Bregman distances expressed in the dual coordinate system induced by the Legendre transformation.}\label{fig:duality}
\end{figure}

Figure~\ref{fig:duality} illustrates the Legendre duality for defining the Riemann-Bregman distances in the dual coordinate systems induced by the Legendre transformation.

\begin{remark}
In information geometry~\cite{AN,CU}, a smooth dissimilarity measure $D$ on a manifold $\cM$ induces a dual structure $(\cM,g_D,\nabla_D,\nabla_D^*)$ defined by a pair of affine connection $(\nabla_D,\nabla_D^*)$ coupled to the metric tensor $g_D$.
The Riemannian geometry $(\cM,g_D)$ is a self-dual information-geometric structure 
with the dual connections coinciding with the Levi-Civita connection $\nabla^{\mathrm{LC}}$. 
Bregman divergences $\delta_\Phi$ yield dually-flat information-geometric spaces (meaning $\nabla_D$-flat and $\nabla_D^*$-flat) although the Riemann-Bregman geometry is usual curved~\cite{N} (meaning $\nabla^{\mathrm{LC}}_D$-curved).
\end{remark}

%%%%%%%%%
\subsection{Examples of Riemann-Bregman distances}\label{dist2}
%%%%%%%%%%

Let us now list a few examples. These will be used below to exemplify the geometric objects defined by the distances and to exemplify the clustering algorithms.

\subsubsection{Case 1: $\phi = x^2/2$}
In this case $\phi''(x)=1$ and $\phi'''(x)=0.$ The induced distance is the standard squared Euclidean distance
$$d_\phi(\bx,\by)^2 = \sum_{i=1}^K (x_i - y_i)^2.$$
\subsubsection{Case 2: $\phi(x)=e^x$}
Now $\phi''(x)=\phi'''(x)=e^x.$ The geodesic distance between $\bx$ and $\by$ is given by
$$d_\phi(\bx,\by)^2 = \sum_{i=1}^K (e^{y_i/2}-e^{x_i/2})^2.$$
\subsubsection{Case 3: $\phi(x)=e^{-x}$} 
The geodesic distance between $\bx$ and $\by$ is given by
$$d_\phi(\bx,\by)^2 = \sum_{i=1}^K (e^{-y_i/2}-e^{-x_i/2})^2.$$
\subsubsection{Case 4: $\phi(x)=x\ln x$ (Shannon negentropy)}
This time our domain is $\cM=(0,\infty)^K$ and $\phi'(x)=(\ln x)-1$, $\phi''(x)=\frac{1}{x}$ and $h(x)=2\sqrt(x)$.
The geodesic distance between $\bx$ and $\by$ is
$$d_\phi(\bx,\by)^2 =2 \sum_{i=1}^K (\sqrt{y_i} - \sqrt{x_i})^2.$$
This looks similar to the squared Hellinger distance used in probability theory. See Pollard's \cite{P} 

\subsubsection{Case 5: $\phi(x)=-\ln x$ (Burg negentropy)}
To finish, we shall consider another example on $\cM=(0,\infty)^K$.
We have $\phi'(x)=-\frac{1}{x}$, $\phi''(x)=\frac{1}{x^2}$ and $h(x)=\log x+c$.
 Now, the distance between $\bx$ and $\by$ is now given by
$$d_\phi(\bx,\by)^2 = \sum_{i=1}^K (\ln y_i - \ln x_i)^2.$$

To continue, let us mention that the thrust in \cite{G} was to examine the prediction process when considering random variables taking values in metric space $(\cM,d_\phi),$ that is to examine the concepts of best predictor of a random variable $\bX$ taking values in $\cM$ when the prediction error is measured in the $d_\phi-$distance. The special relation of the distance (\ref{dist}) to the Euclidean distance makes the following result clear:

\begin{theorem}[Separable Riemann-Bregman centroid]\label{sampmean}
Given a collection of points $\{\bx_1,...,\bx_N\}$ in $\cM,$ the point $\hat{\bx}$ the realizes the minimum of
$$\sum_{n=1}^N d_\phi(\bx_n,\bxi)^2\;\;\;\mbox{over}\;\;\;\bxi\in\cM$$
is unique and given by
\begin{equation} \label{eq:center} 
\hat{\bx} = H\Big(\frac{1}{N}\sum_{n=1}^N h(\bx_n)\Big).
\end{equation}
\end{theorem}

Observe that formula Eq.~\ref{eq:center} coincides with the left-sided Bregman centroid~\cite{N3}.
However, left-sided Bregman $k$-means and Riemann-Bregman $k$-means will differ in the assignment steps because they rely on different dissimilarity measures.

We shall make extensive use of this definition when computing the centers of the clusters in the various metrics. This is done in Section 4 below, where we start from a given data set and organize it in clusters according to different clustering procedures for each of the distances, and compute the centers of the resulting clusters each distance as in (\ref{eq:center}).

In Section 2 we consider some standard geometric objects, like Balls, Voronoi cells, in the different distances, whereas in Section 3 we consider the problem of fixed rate quantization for $\cM-$valued random variables .

%%%
\section{Balls and Voronoi cells in the derived metrics}
%%%%
\subsection{Balls}
The examples listed show that $h:(\cM,d_\phi) \to (\mathbb{R}^K,\|\cdot\|)$ is a non-linear isometry. This allows us to
describe balls in $(\cM,d_\phi)$ as
\begin{equation}\label{ball}
B_\phi(\bx,r) = \{\by\in\cM: d_\phi(\by,\bx)\leq r\} = h^{-1}\Big(B(h(\bx),r)\Big).
\end{equation}
Here,  $B(h(\bx),r)$ denotes the ball in $\mathbb{R}^K,$ of radius $r$ centered at $h(\bx)$.
The space of balls can also be defined using a potential function following~\cite{BNN}.
 
\subsection{Voronoi diagrams}

\begin{figure} 
 \centering
	 \subfloat[$\exp(x)$]{
                \includegraphics[width=2.0in,height=2.0in]{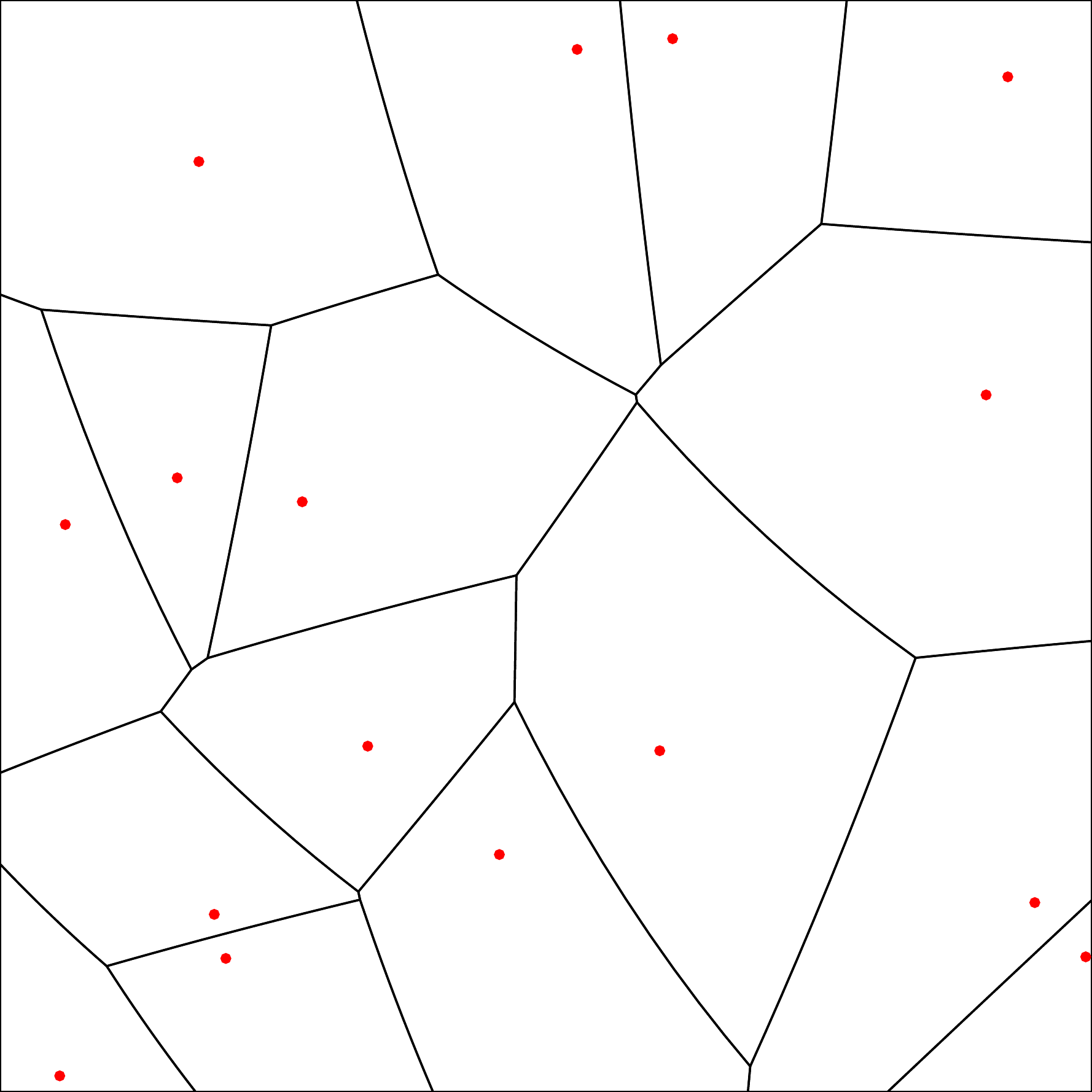}
              }
        \subfloat[$\exp(-0.5x)$]{
                \includegraphics[width=2.0in,height=2.0in]{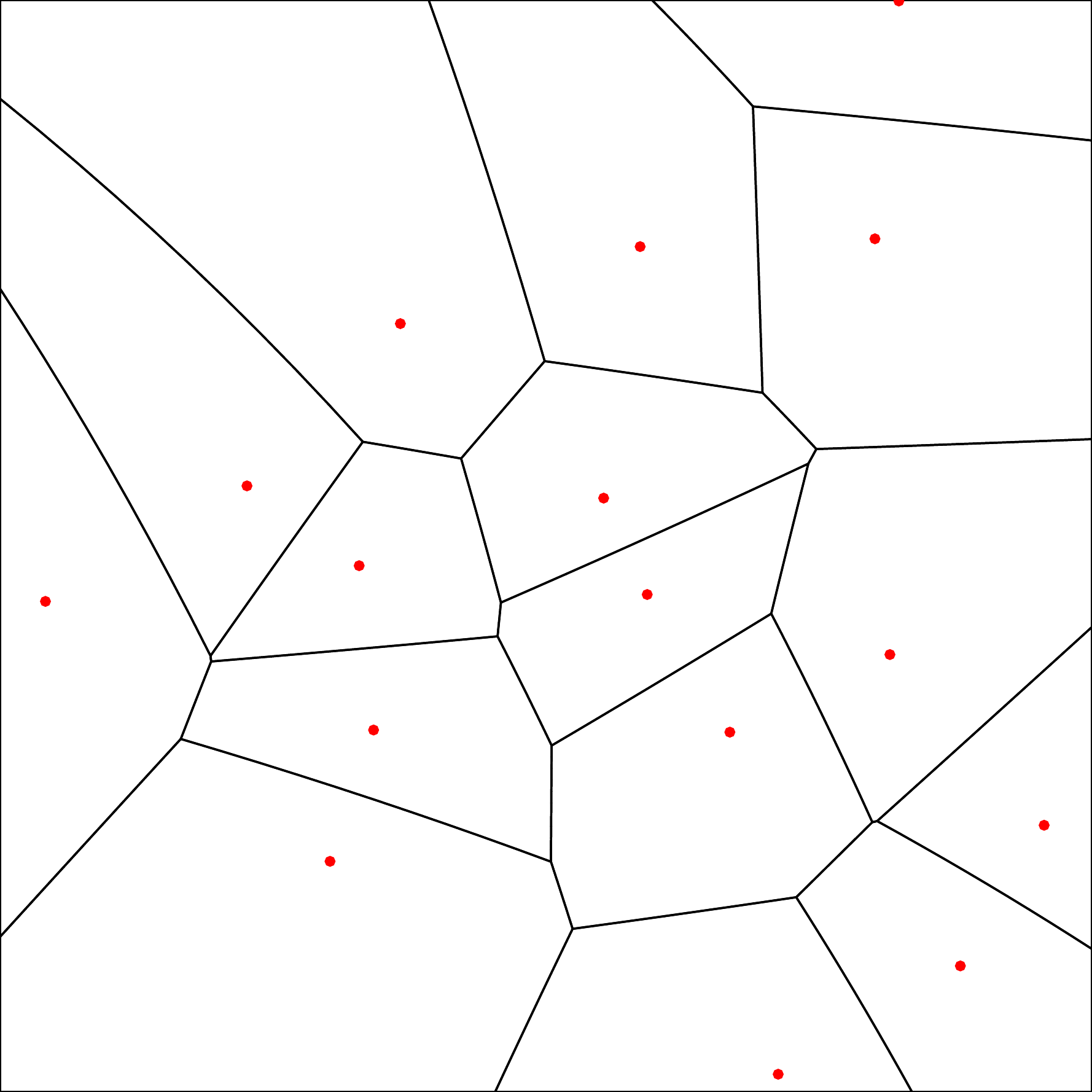}
                }
	\subfloat[$-x\ln x$]{
                \includegraphics[width=2.0in,height=2.0in]{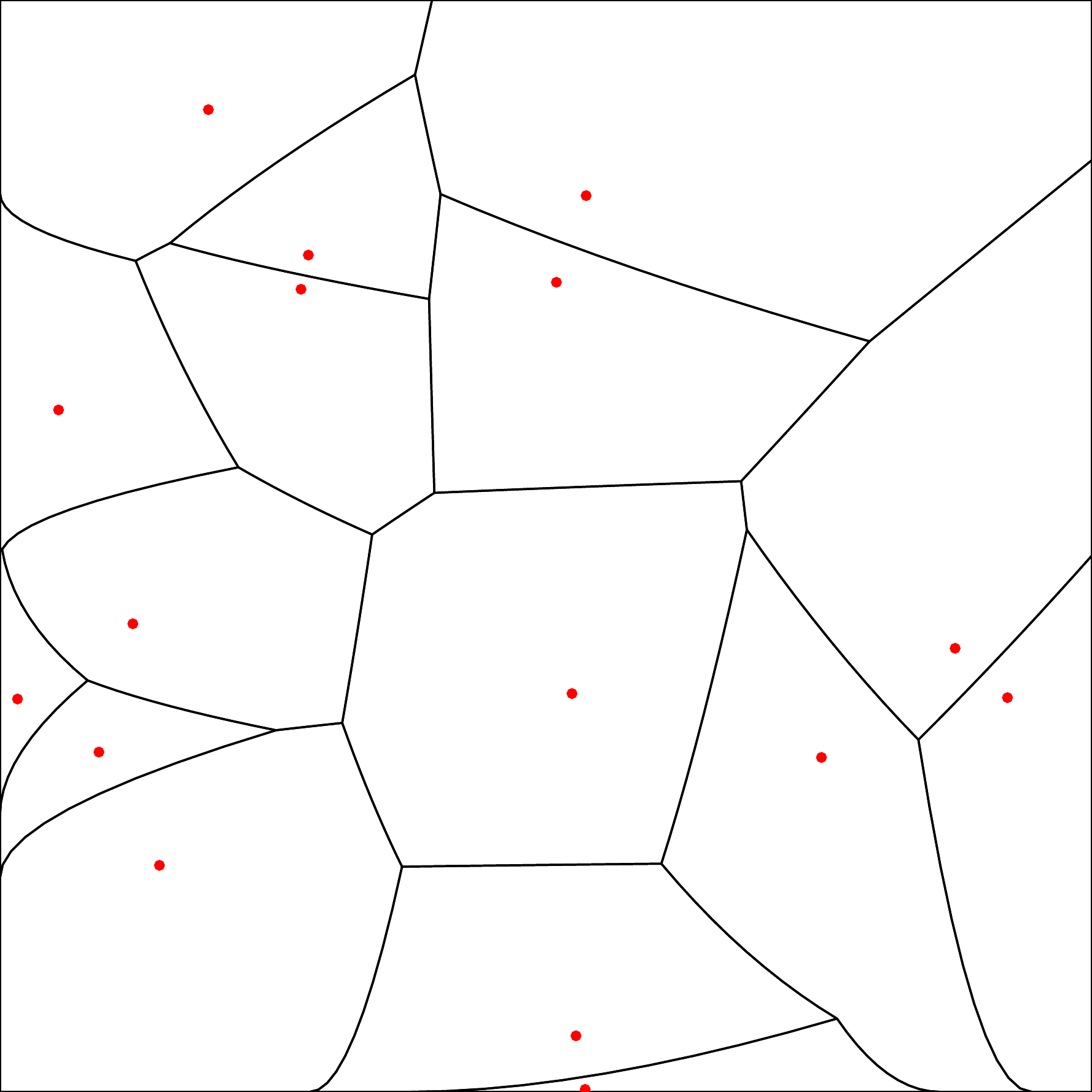}
                 }
\caption{Voronoi diagrams induced by the Riemann-Bregman distances}   
\label{vorcell}   
\end{figure}

Let $\cS=\{\by_1,...,\by_n\}$ be some subset $(\cM,d_\phi).$ The Voronoi cell in $\cM$ of $\by_i$ is defined by
 \begin{equation}\label{vor}
\mathrm{vor}_\phi(\by_i) = \{\bx \in \cM\ :\ d_\phi(\bx,\by_i) \leq d_\phi(\bx,\by_j),\;\forall \by_j\in \cM\}.
\end{equation}
And as above, if we denote by $\mathrm{vor}(h(\by_i))$ the Voronoi cell of $h(\by_i)$ in the Euclidean metric, it is clear that
\begin{equation}\label{vor2}
\mathrm{vor}_\phi(\by_i) = h^{-1}\big(\mathrm{vor}(h(\by_i))\big).
\end{equation}
A few graphical examples are shown in Figure \ref{vorcell}.

\begin{figure}
\centering
\subfloat[$x\ln x$ (Shannon)]{
                \fbox{\includegraphics[width=0.4\textwidth]{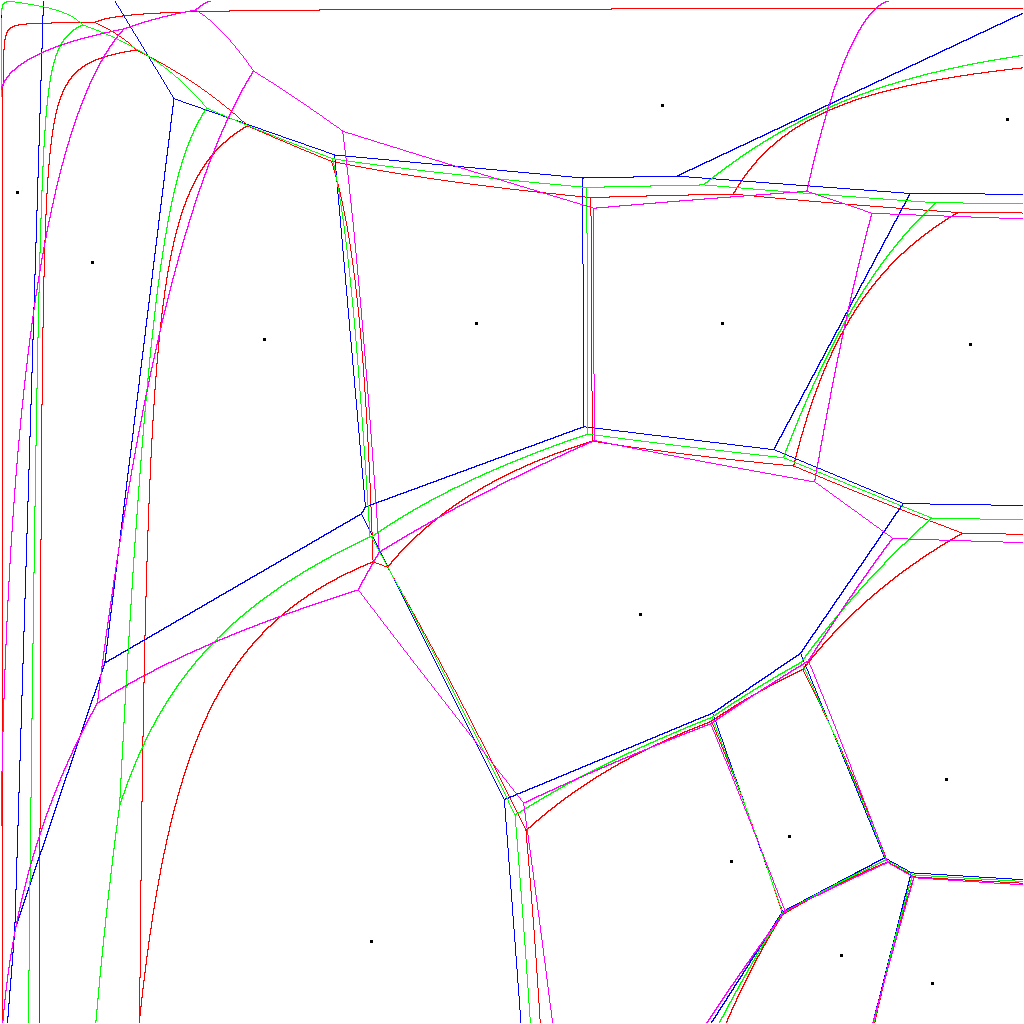}}
              }
							\hskip 1cm
        \subfloat[$-\ln x$ (Burg)]{
                \fbox{\includegraphics[width=0.4\textwidth]{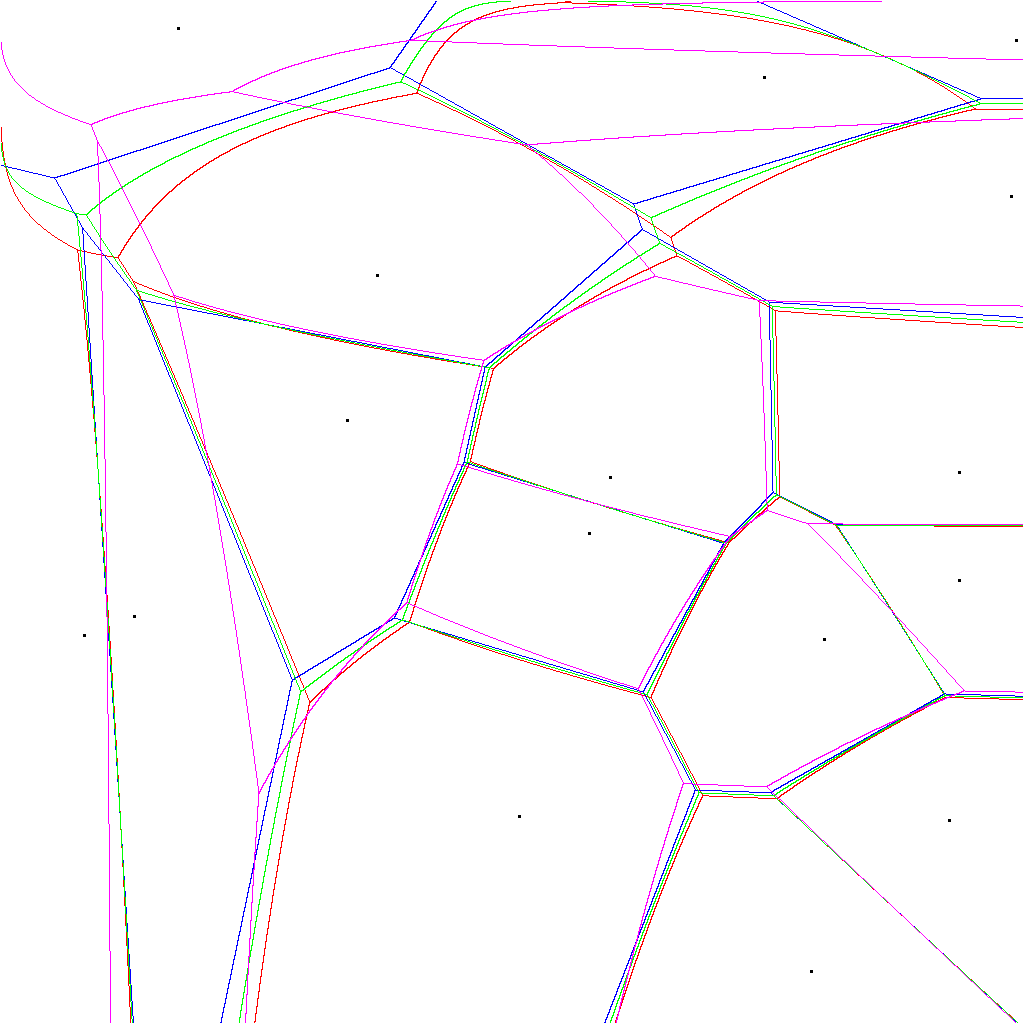}}
                }
\caption{Four types of Voronoi diagrams induced by a Bregman generator: The left-sided Bregman Voronoi diagram (blue), the right-sided Bregman Voronoi diagram (red), the symmetrized Bregman Voronoi diagram (green), and the Riemann-Bregman Voronoi diagram (purple).}   
\label{dvorcell}
\end{figure}

Figure~\ref{dvorcell} displays four types of Voronoi diagrams induced by a Bregman generator;
The pictures illustrate the fact that the symmetrized Bregman Voronoi diagram is different from the (symmetric metric)  Riemann-Bregman Voronoi diagram.

\section{Fixed rate quantization}
Let $\cM \subseteq \mathbb{R}^K$ be provided with a metric $d_\phi$ derived from a divergence. Let $N\geq 1$ be a given integer and $\cC=\{y_1,...,y_N\}\subset\cM$ be a given set, called the {\it codebook} and its elements are called the {\it code vectors}. A fixed rate quantizer (with codebook $\cC$) is a measurable mapping $q:\cM\to\cC.$ 
The cardinality of $\cC$ is called the quantization rate. Two quantizers have the same rate if their codebooks have the same cardinality

For a $\cM-$valued random variable $\bX$ defined on a probability space $(\Omega,\
\cF,P),$ the $\phi-$distortion of $q$ in quantizing $\bX$ is given by the expected reconstruction square error:

\begin{equation}\label{disterr}
D_\phi(\bX,q(\bX))^2 = E[d_\phi(\bX,q(\bX))^2].
\end{equation}

If we are interested in only one $\cM-$valued random variable $X,$ we may replace $\Omega$ by $\cM,$ think of $\cF$ as the Borel sets of $\cM,$  think of $\bX$ as the identity mapping, and think of $P$ as a probability measure on $(\cM,\cF).$ Also, our notation is slightly different from that of Linder \cite{Lin} in that we put a square into the definition of the square error. Since $q(\bX)$ is constant on the ``level'' sets $S_i:=\{q(X)=\by_i\}$ for $i=1,...,N,$ the former expectation can be written as
$$D_\phi(\bX,q(\bX))^2 = E[E[d_\phi(\bX,q(\bX))^2|\sigma(q)]].$$
Here we denote by $\sigma(q)$ the sub$-\sigma-$algebra of $\cF$ generated by the partition $\{S_1,...,S_N\}$ of $\Omega$ induced by the quantizer $q.$ When $P(S_i)>0$ holds for all blocks of the partition, the conditional expectation takes values
\begin{equation}\label{condquat1}
E[d_\phi(\bX,q(\bX))^2|\sigma(q)] = \frac{1}{P(S_i)}\int_{S_i} d_\phi(\bX,\by_i)^2dP\;\;\; \mbox{on the set}\;\;S_i, i=1,...,N.
\end{equation}
Notice that given $\sigma(q),$ there is a quantizer having the same rate and realizing an error smaller than $D_\phi(\bX,q(\bX))^2.$ The existence and uniqueness of such quantizer is given by the following result in \cite{G}.

\begin{theorem}\label{condexp1}
 Let $\cG$ be a sub$-\sigma-$algebra of $\cF.$ Let $\bX$ be such that $h(\bX)\in\cL_2$ (Lebesgue space of functions that are square integrable).  Then there exists a unique $\cG$ measurable random variable, denoted by $E_\phi[X|\cG]$ satisfying
$$E_\phi[X|\cG] = \textrm{argmin} \{D_\phi(\bX,\bY)^2| \bY\in\cG,\;h(\bY)\in \cL_2\}$$
\end{theorem}
The result mentioned a few lines above can be stated as
\begin{theorem}\label{optquant1}
Let $q$ be a quantizer of rate $N.$ Suppose that the blocks $\{S_i:i=1,...,N\}$ of the partition of $\Omega$ determined by $q$  are such that $P(S_i)>0$ for all $i=1,...,N.$ Then 
\begin{equation}\label{optcb1}
q^*(\bX) = E_\phi[X|\cG] = \sum_{i=1}^N\frac{1}{P(S_i)}E_\phi[\bX;S_i]I_{S_i}(\bX) := \sum_{i=1}^N\by_iI_{S_i}(\bX).
\end{equation}
is the quantizer with codebook $\cC^*,$  whose elements are $\by^*_i=\frac{1}{P(S_i)}E_\phi[\bX;S_i],$ which makes the error $D_\phi(\bX,q(\bX))^2$ smaller over all codebooks which determine the same partition of $\Omega.$
\end{theorem}

An interesting way to associate sets to a codebook $\cC=\{\by_1,...,\by_N\}$ of rate $N$ is by means of Voronoi cells. $\{S_1,...,S_N\}$ are the Voronoi cells determined by $\cC,$ their intersections are polyhedra of dimension lower than $K$ (faces), and the probability $P$ does not change them, that is if $P(S_i\cap S_j)=0$ whenever $i\not= j.$ If we put $B_i=\textrm{int}(S_i)$ for the interior of the Voronoi cells, they generate a sub-$\sigma-$algebra of $\cF$ such that $\cup_iB_i$ differs from $\cM$ by a set of $P-$probability $0.$

In the notation of Theorem \ref{optquant1} we have
\begin{corollary}\label{coro1}
If the $\sigma-$algebra $\cG$ is generated buy the interiors of the Voronoi cells determined by a codebook $\cC,$
then
$$\by^*_i = \frac{1}{P(S_i)}E_\phi[\bX;S_i]  = \frac{1}{P(B_i)}E_\phi[\bX;B_i]\in B_i\subset S_i,\;\;\;\forall i=1,...,N.$$
\end{corollary}
The proof is clear: The integral is a weighted average of points in a convex set.\\

%%%%%%%
\section{Riemann-Bregman Clustering}
%%%%%%
In this section we shall consider a $2-$dimensional data set in which several clusters (indicated) by different colors in the figures displayed below are apparent.  We shall suppose that the base space in which the random variable takes values comes equipped with one of the five distances listed in Section \ref{dist2}. 

We shall suppose that the points belong one of the possible manifold on which a Riemann-Bregman metric has been defined, and the object of the exercise is to identify the clusters. To assign the data points into clusters, we shall apply the $k$-means, the EM (Expectation-Maximization), the HCPC (Hierarchical Clustering Principal Component) and the HAC (Hierarchical Agglomerative Clustering) techniques. Once the clusters are identified, we determine their centers by invoking Theorem (\ref{sampmean} and (\ref{eq:center}). In the four examples described below, we shall see how the choice of the metric determines the clusters.\\

The points were generated all in $(0,\infty)^2$ in order to be able to visually compare the result of the clustering procedures in the distances listed on Section \ref{dist2}.

Strategy: \\
{\bf Step 1} Generate a cloud $\{\bx_1,....,\bx_n\}$  in  $(0,\infty)^2,$\\
{\bf Step 2} Apply the $h-$ mapping and form the cloud $\{h(\bx_1),...,h(\bx_N)\}$ in $h(\cM).$ \\
{\bf Step 3} Apply the clustering algorithms. If $\{\bet_1,...,\bet_M\}$ are the clusters of the new cloud, then the clusters in the original coordinates are $\{\by_1=h^{-1}(\bet_1), ..., \by_n=h^{-1}(\bet_n)\}$.\\
{\bf Step 4} Plot the ``centers'' $\{\by_1,...,\by_n\}$ in the original cloud.\\

{\bf Notice that in all examples, the mappings $h$  and $h^{-1}$ are invertible, and are applied componentwise.}
The aim of the exercise is to see how the choice of the metric determines the clusters.\\
In particular, we observe that although the centroid update rule coincide with the left-sided Bregman clustering~\cite{N3}, the assignment steps of the left-sided Bregman clustering and the Riemann-Bregman clustering are different.

\subsection{The data set}
For all the the numerical experiments considered below, the ``original'' data is composed of samples of four pairs of independent normal variables with parameters specified in Table \ref{err} shown next.
		\begin{table} 
			\centering%\rowcolors{4}{gray!6}{white}
						\begin{tabular}{ccccc}
				\hiderowcolors
				\hline 
				Cluster            &     $$       & $X_1$ & $X_2$  &\\  \hline \hline
				\showrowcolors
				\multirow{2}{*}{1} & $\sigma_1$    &  0.5  &  0.5  &\\
				                   & $\mu_1$       &   8   & 6.5   &\\  \hline 
			 	\multirow{2}{*}{2} & $\sigma_2$    &   0.4 & 0.45  &\\
				                   & $\mu_2$       &   9   & 7.5   &\\  \hline 
				\multirow{2}{*}{3} & $\sigma_3$    &  0.35 & 0.35  & \\
			  	                   &  $\mu_3$      &  8.5  & 9     &\\  \hline 
				\multirow{2}{*}{4} & $\sigma_4$    &  0.6  & 0.6   & \\
			                       &  $\mu_4$      &  8    &  10   & \\  \hline 
			\end{tabular}
			\rowcolors{2}{white}{white}
			\caption{Centers of $k$-means clusters for a groundtruth synthetic dataset.}
			\label{err}
		\end{table}

Thus we obtain four point clouds with some overlap. The number of points in the four original groups is, respectively, $100, 300, 200, 150.$ These numbers will be used as reference to check upon the performance of each clustering technique when the distance between points is measured in each of the five different Riemann-Bregman metrics.

\subsection{Clustering by $k$-means}
The k-means clustering algorithm is an iterative methodology that assigns data points to a cluster, in such a way  that the sum of the squared distances between the data points and the cluster’s centroid (defined as the arithmetic mean) is minimal. The less variation we have within clusters, the more homogeneous the data points are within the same cluster. 

\begin{figure} 
 \centering
	 \subfloat[Data set\label{dataset1a}]{
                \includegraphics[width=2.0in,height=2.0in]{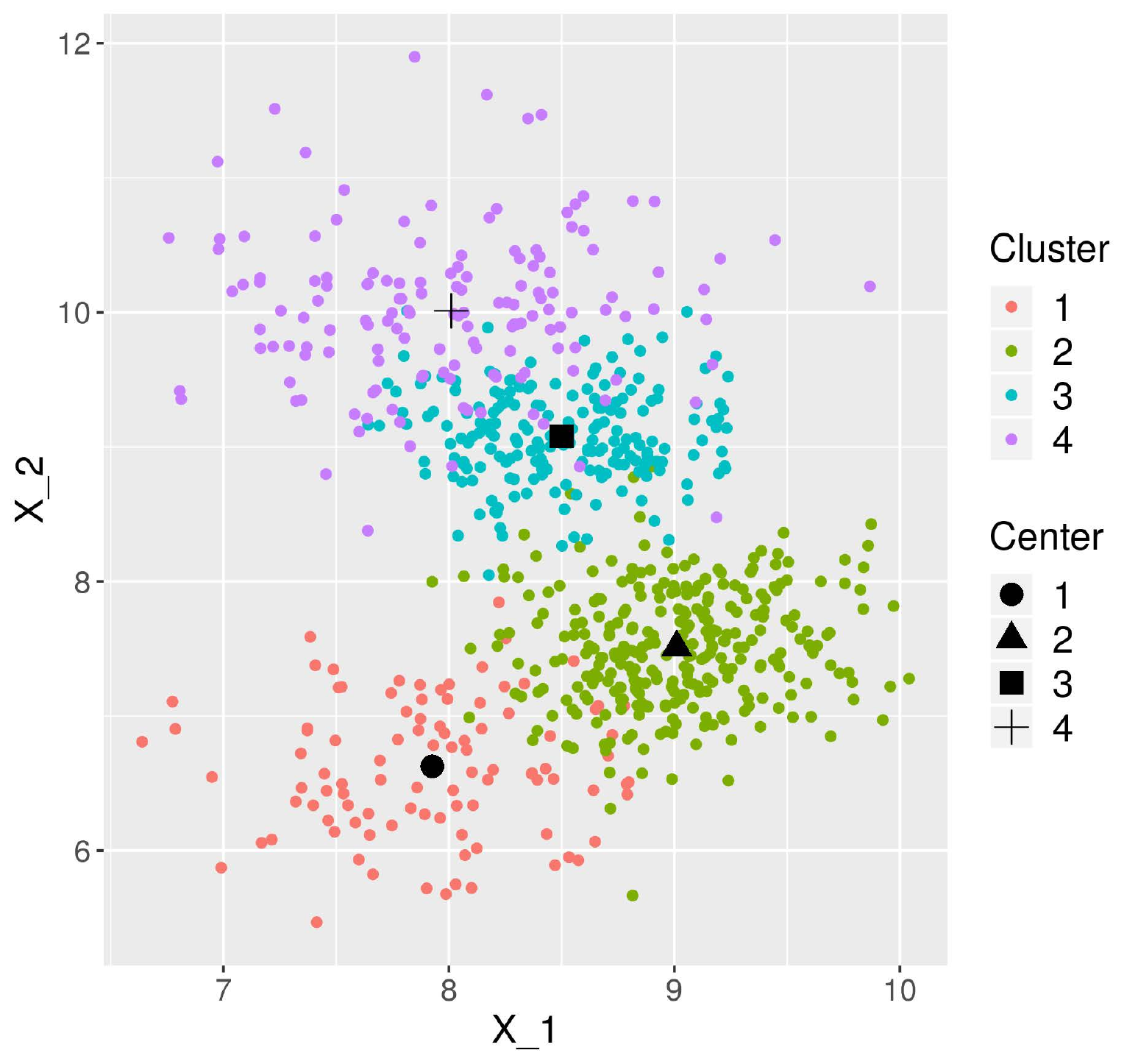}
               }
    \subfloat[$h(x)=x$]{
                \includegraphics[width=2.0in,height=2.0in]{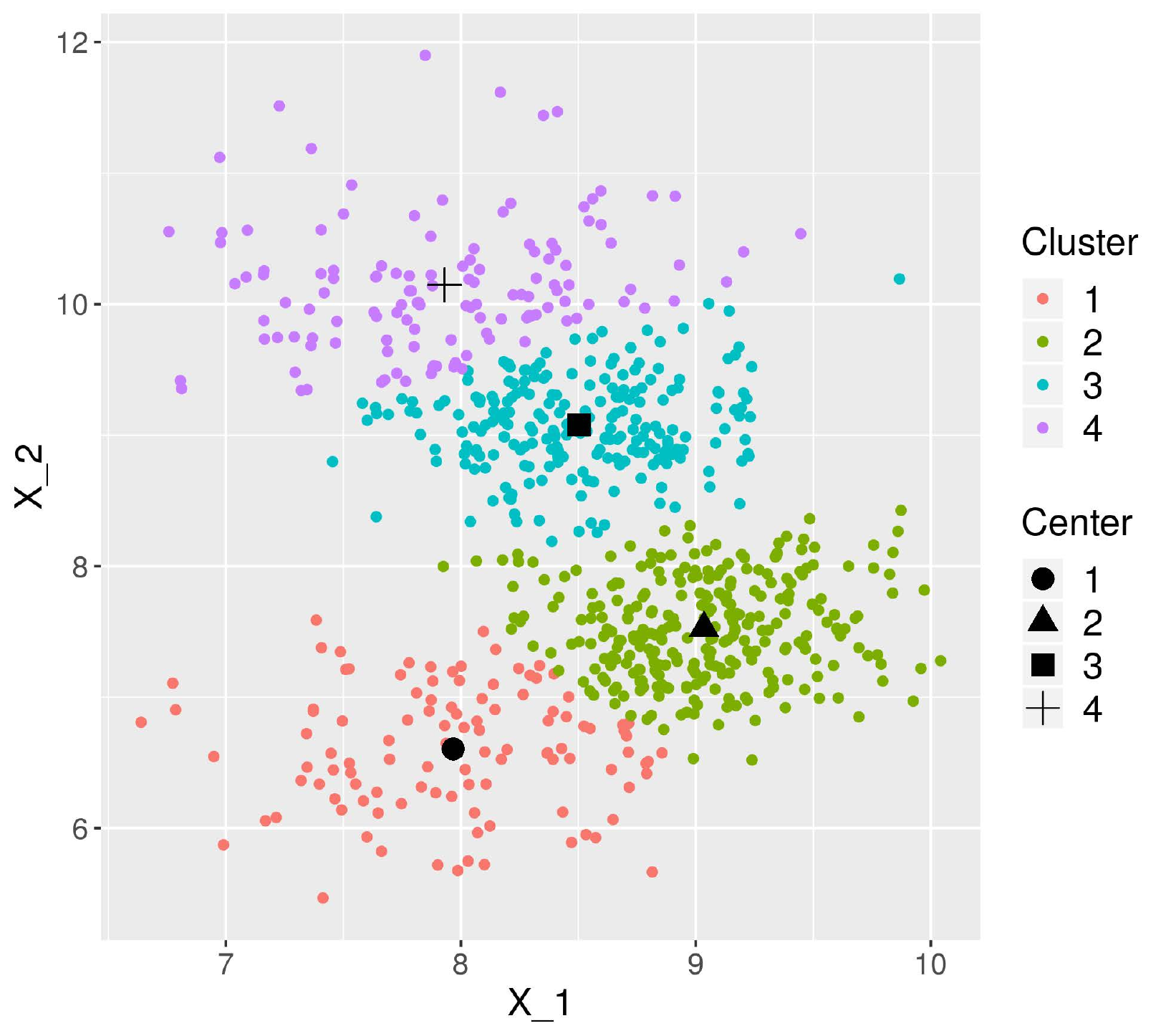}
                 }
	\subfloat[$h(x)=\exp(-0.5x)$]{
                \includegraphics[width=2.0in,height=2.0in]{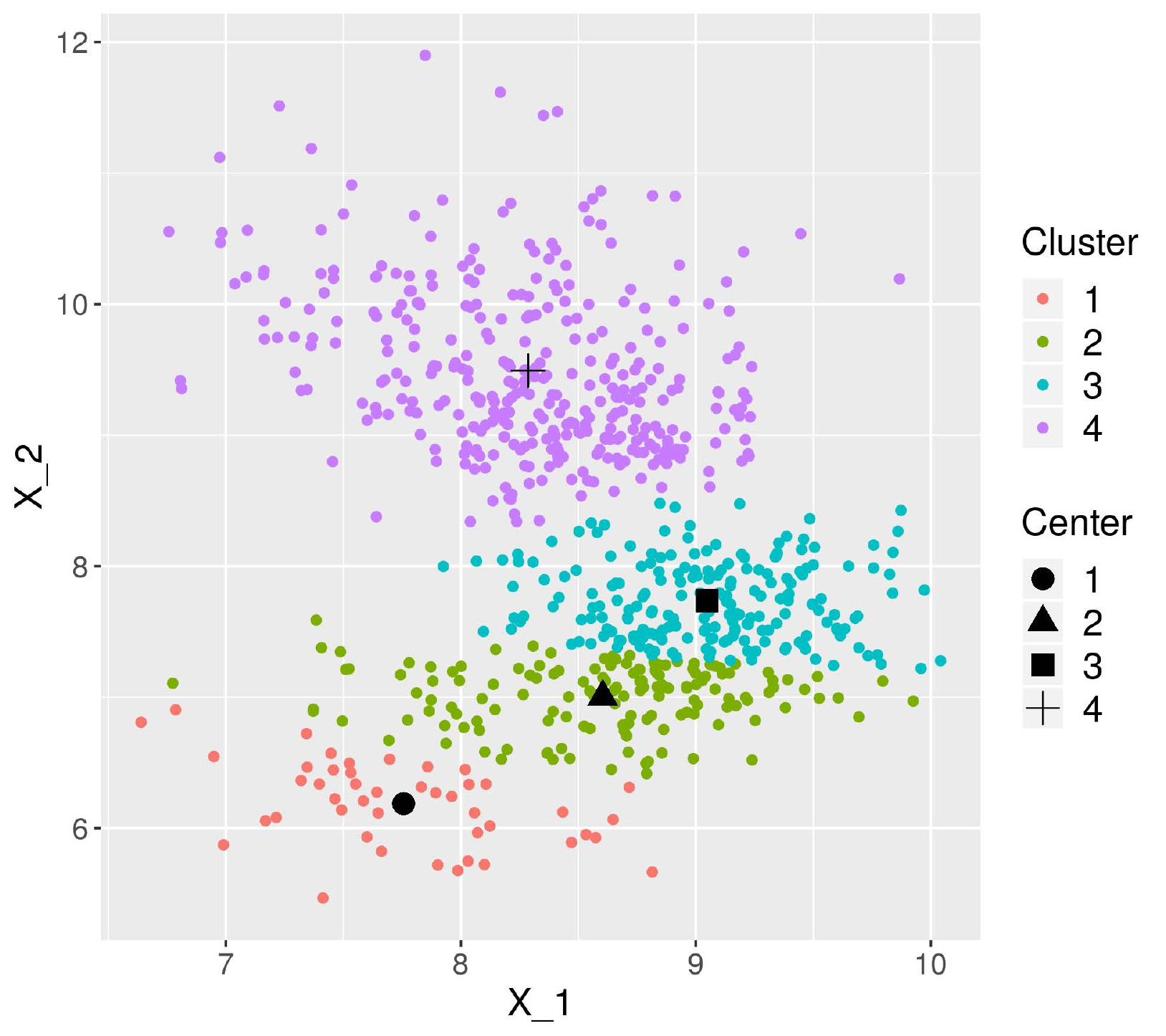}
                 }
								
  \subfloat[$h(x)=\exp(0.5x)$]{
                \includegraphics[width=2.0in,height=2.0in]{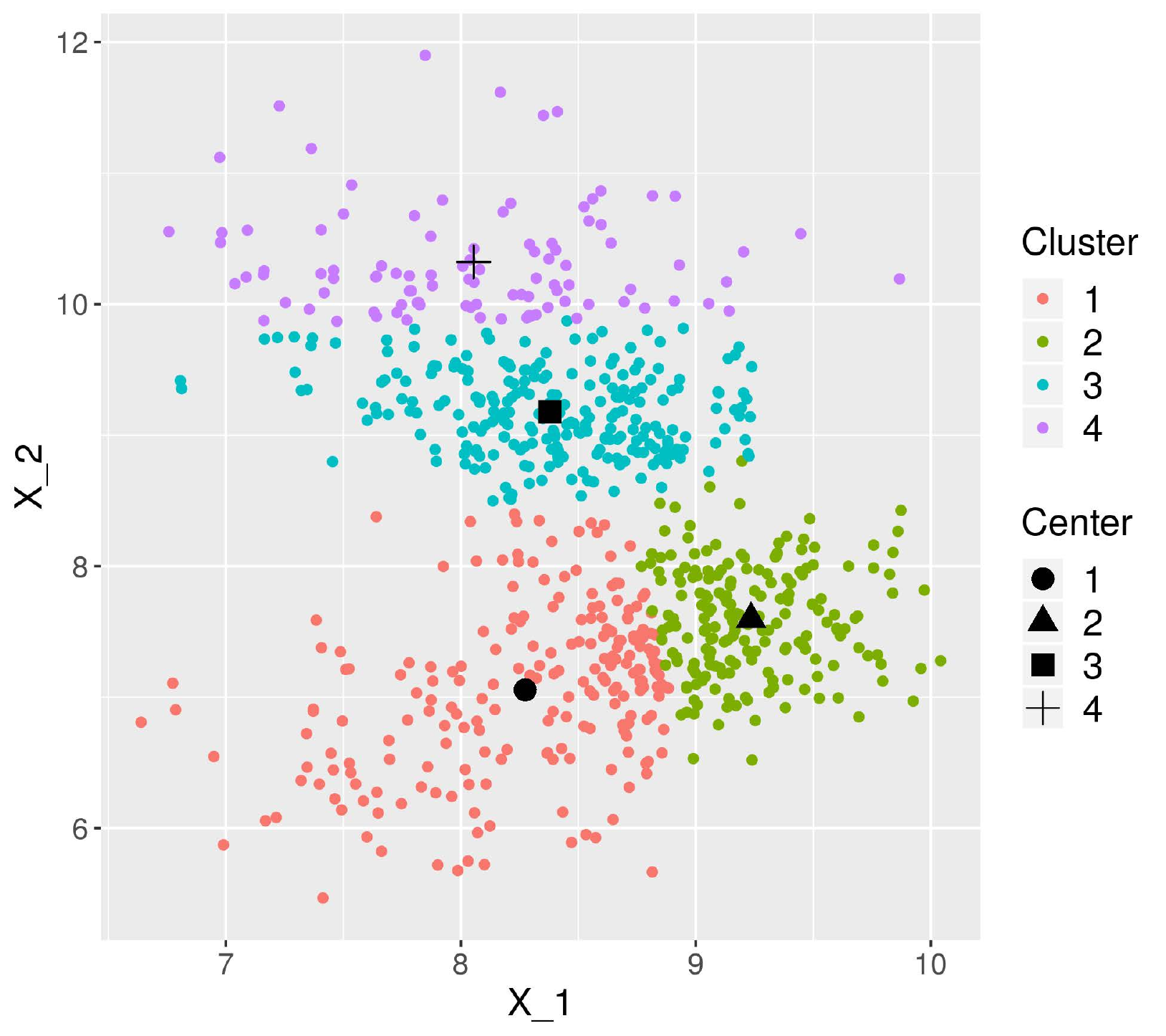}
               }
    \subfloat[$h(x)=\sqrt{x}$]{
                \includegraphics[width=2.0in,height=2.0in]{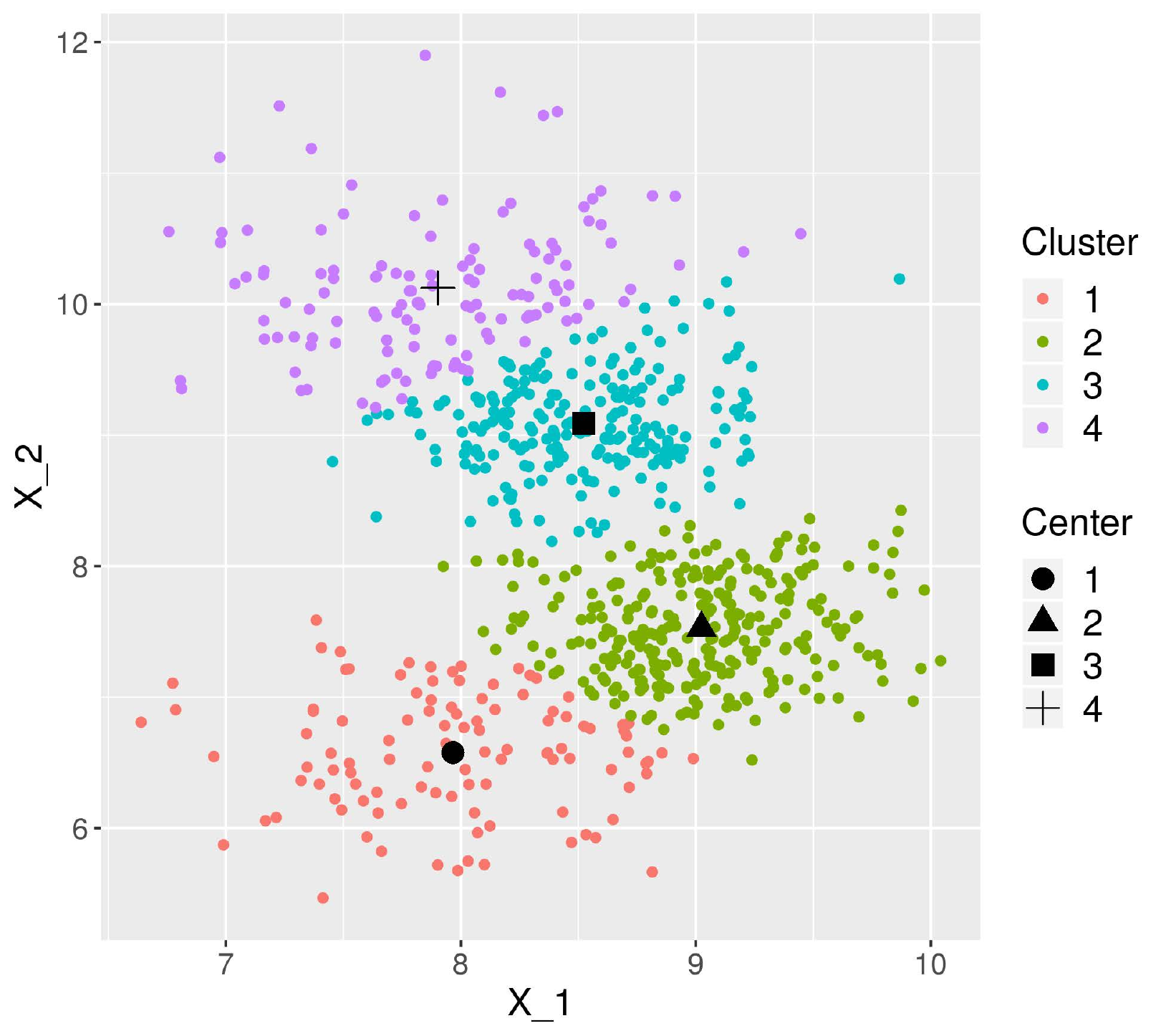}
                 }						
			   \subfloat[$h(x)=\ln x$]{
                \includegraphics[width=2.0in,height=2.0in]{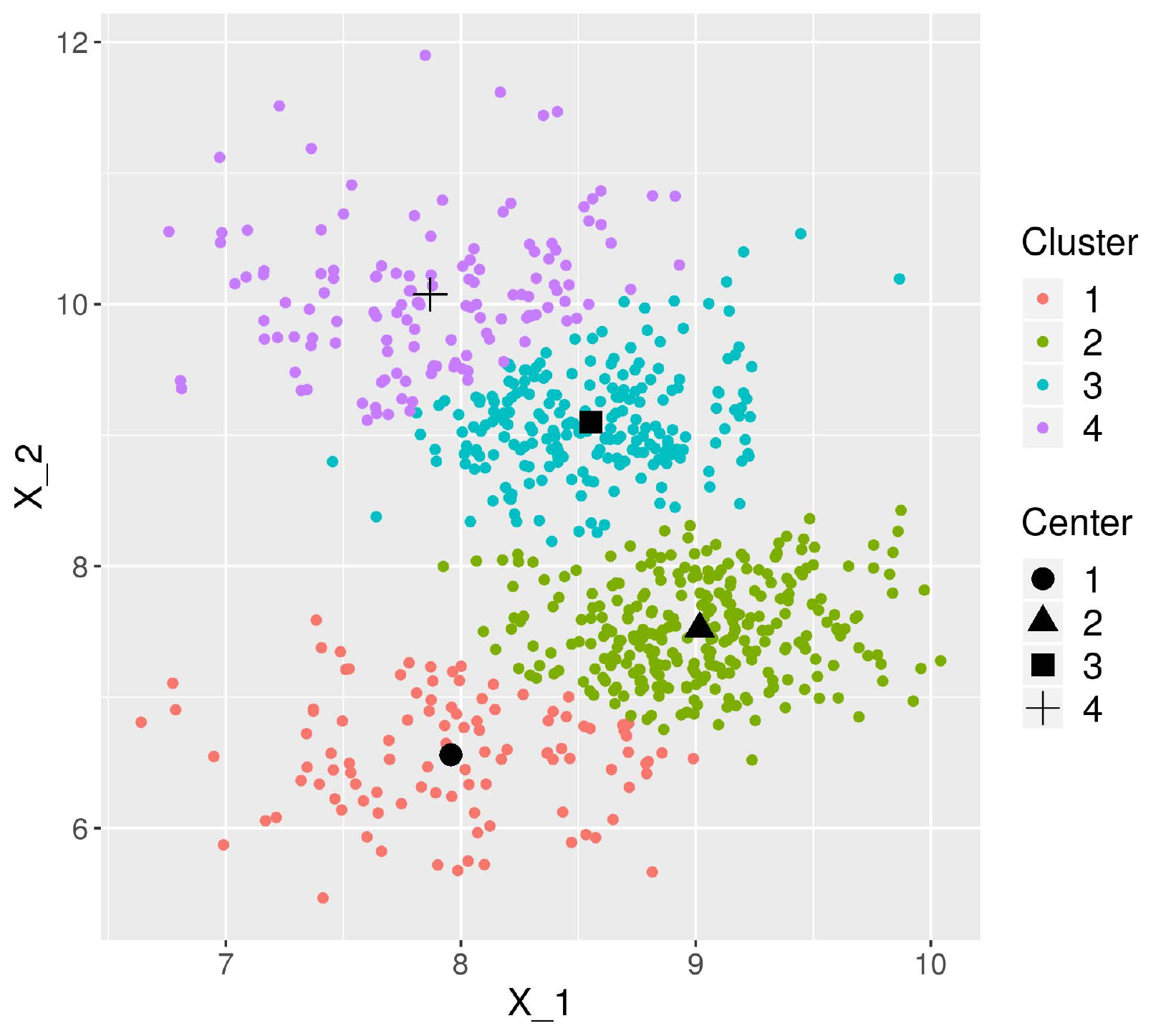}
                 }					
\caption{Clustering by $k$-means}   
\label{kclusters}   
\end{figure}

In Figure \ref{kclusters} we present the result of the application of the $k-$means procedure to the original data set depicted in panel \ref{dataset1a}. The caption of each of the panels describes the $h-$function that has been applied to the data before applying the clustering procedure. Once the clusters are determined, the in-cluster mean is computed and the whole set is mapped back to the original space. 

We add at this point that if one suspects (or knows) that the data sets are in different scales, it might be convenient to standardize the data. This is important since clustering algorithms are based on distance comparison procedures.
In~\cite{BBC2}, Banerjee et al. highlight the bijection between regular exponential families and a class of Bregman divergences called regular Bregman divergences. 
They demonstrate experimentally that the $k$-means clustering perform best in practice on mixture of exponential families (with cumulant function $\phi$) when the Bregman divergence is chosen for the dual convex conjugate $\phi^*$.

The centers of the original data set and those of the clusters are listed in Table \ref{tab:center}.

	\begin{table} 
	\small
	\begin{tabular}{ccccc}		 \hline 
		Cluster & 1 & 2 & 3 & 4\\  \hline \hline
		original & (7.93, 6.63) & (9.01, 7.51) & (8.5, 9.08) & (8.01, 10.01)\\
		$h(x) = x$ & (7.97, 6.6) & (9.04, 7.53) & (8.5, 9.08) & (7.93, 10.15)\\
		$h(x) = \sqrt{x}$ & (7.97, 6.58) & (9.02, 7.53) & (8.52, 9.09) & (7.9, 10.12)\\
		$h(x) = \ln(x)$ & (7.96, 6.56) & (9.02, 7.52) & (8.55, 9.1) & (7.87, 10.07)\\
		$h(x) = \exp(0.5x)$ & (8.27, 7.06) & (9.24, 7.6) & (8.38, 9.18) & (8.05, 10.32)\\
		$h(x) = \exp(-0.5x)$ & (7.76, 6.19) & (8.6, 7) & (9.05, 7.74) & (8.29, 9.49)\\
	\end{tabular}
	\rowcolors{2}{white}{white}
		\caption{centers of $k$-means clusters obtained with different methodologies\label{tab:center}}
	 
\end{table}

From the table it is clear that the Euclidean distance and the logarithmic distance perform better than the others in capturing the original centers of the data. Actually, the centers of the original data are the minimizers to the original data set in the Euclidean distance and are the empirical means of the normals used to generate the data. 

As a further test of the clustering algorithm we provide a ``headcount'' of points in each cluster and compare the numbers with the number of original points in each cluster. The result is displayed in Table \ref{size}.

\begin{table} 
	\centering\rowcolors{2}{gray!6}{white}	
	\begin{tabular}{cccccc}
		\hiderowcolors
		\hline
		Cluster & 1 & 2 & 3 & 4 & \textbf{Accuracy} \\ \hline \hline  
				\showrowcolors
		Original            & 100 & 300 & 200 & 150 & 1\\
		$h(x) = x$          & 109 & 286 & 226 & 129 & 0.953\\
		$h(x) = \sqrt{x}$   & 106 & 289 & 225 & 130 & 0.959\\
		$h(x) = \ln(x)$      & 103 & 292 & 221 & 134 & 0.968\\
		$h(x) = \exp(0.5x)$  & 214 & 196 & 240 & 100 & 0.768\\
		$h(x) = \exp(-0.5x)$ & 46  & 146 & 211 & 347 & 0.388\\
	\end{tabular}
	\rowcolors{2}{white}{white}
    \caption{Size of the $k$-means clusters and accuracy obtained with different transformations}
	\label{size}
\end{table}

%%%%
\subsection{Clustering with the EM algorithm}
%%%%
The  expectation maximization (EM) algorithm~\cite{EM} is an iterative method to estimate the parameters in a  statistical model, in which the model depends on unobserved variables. The EM iteration alternates between performing an expectation (E) step and a maximization (M) step. In the first step a likelihood function is created and in the second it is maximized. See \cite{MP} for details.

In the EM methodology an initial guess is made for the model’s parameters and a distribution is created, this is the E-step or Expected distribution, then the parameters are updated in the (M-step) maximizing the expectation computed in the E-step. Both steps are repeated until the convergence is clear (distribution that doesn’t change from the E-step to the M-step)

\begin{figure} 
 \centering
	 \subfloat[Data set]{
                \includegraphics[width=2.0in,height=2.0in]{originalclusters.pdf}
               }
    \subfloat[$h(x)=x$]{
                \includegraphics[width=2.0in,height=2.0in]{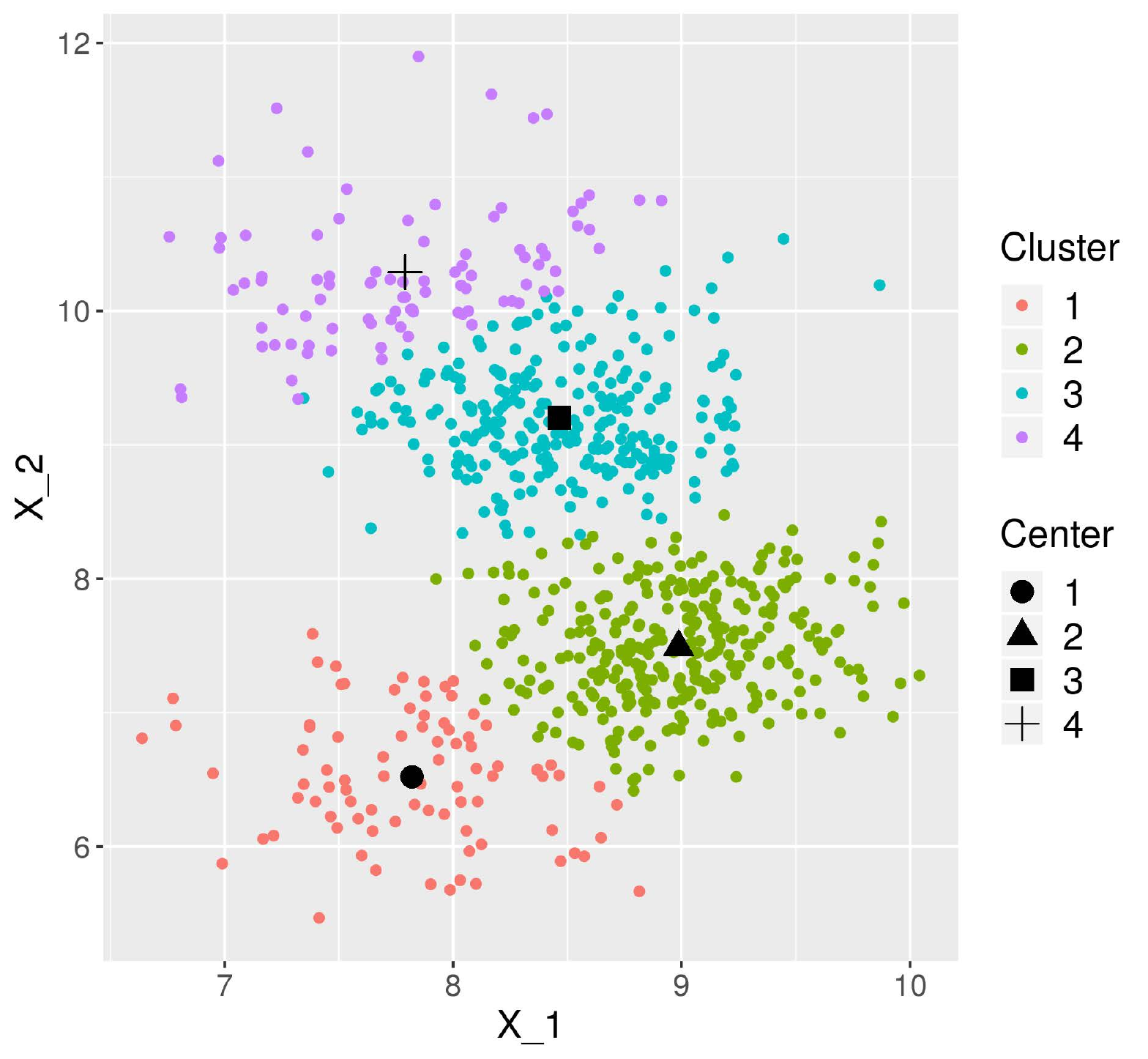}
                 }
	\subfloat[$h(x)=\exp(-0.5x)$]{
                \includegraphics[width=2.0in,height=2.0in]{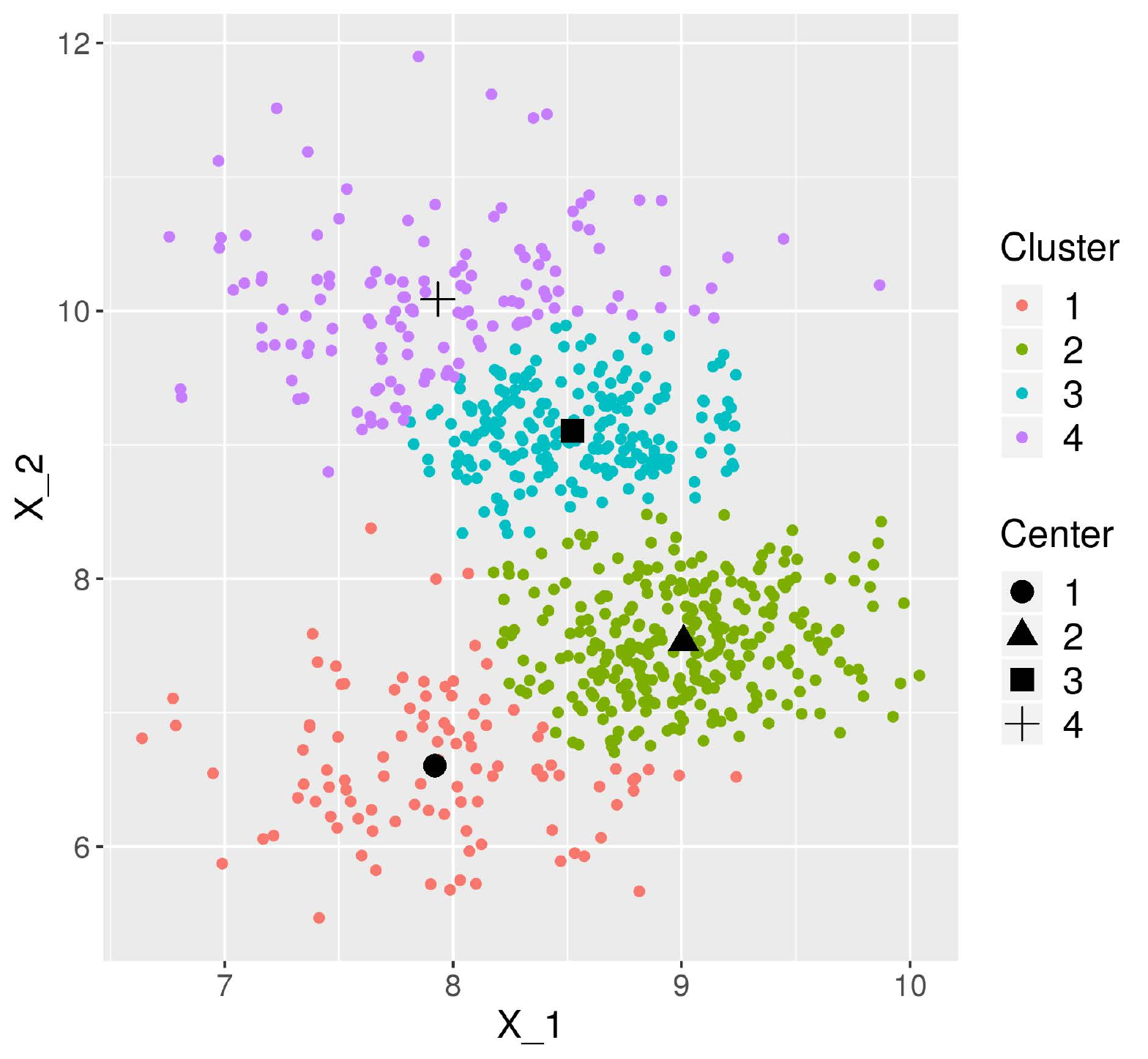}
                 }
								
  \subfloat[$h(x)=\exp(0.5x)$]{
                \includegraphics[width=2.0in,height=2.0in]{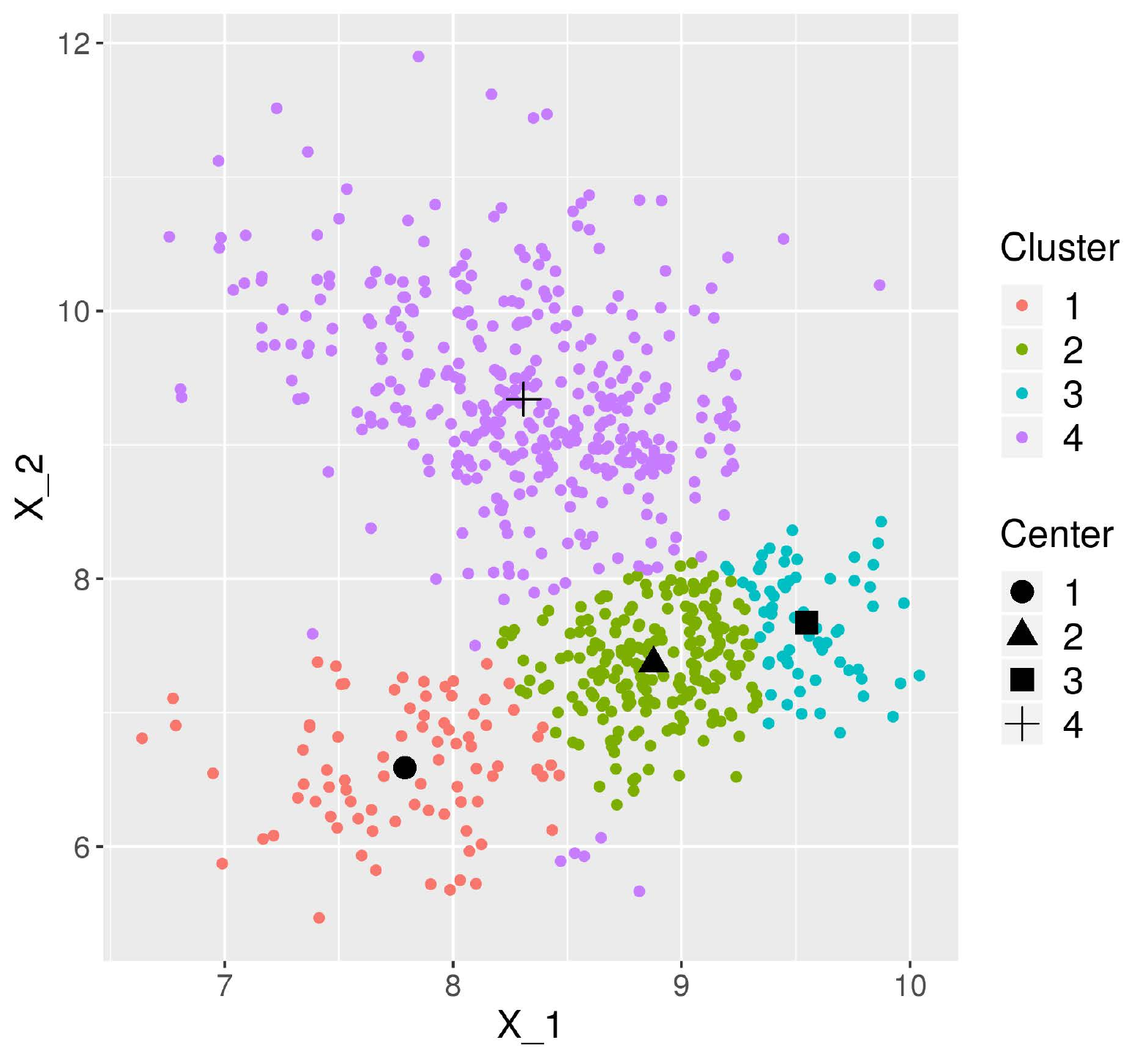}
               }
    \subfloat[$h(x)=\sqrt{x}$]{
                \includegraphics[width=2.0in,height=2.0in]{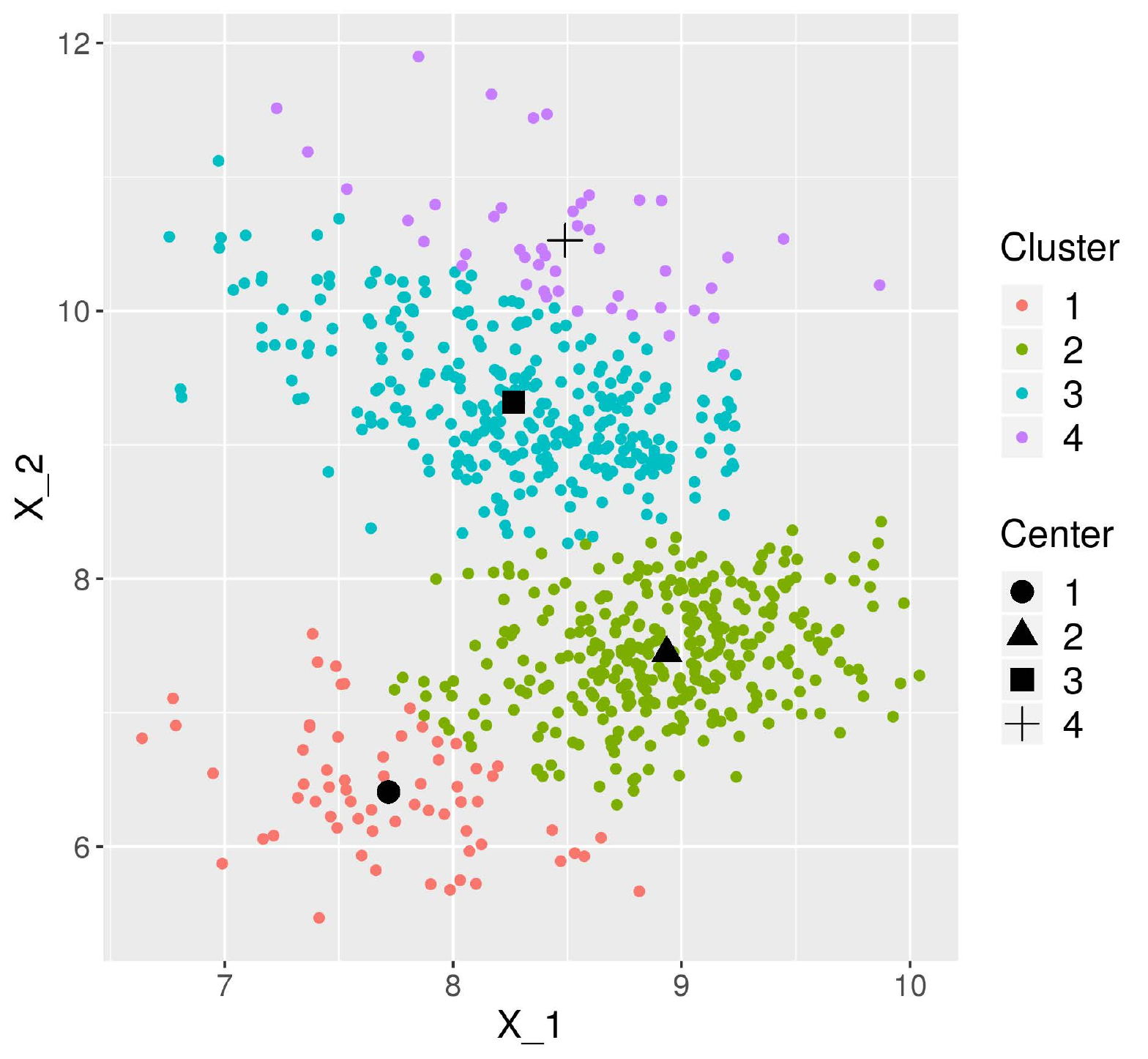}
                 }						
			   \subfloat[$h(x)=\ln x$]{
                \includegraphics[width=2.0in,height=2.0in]{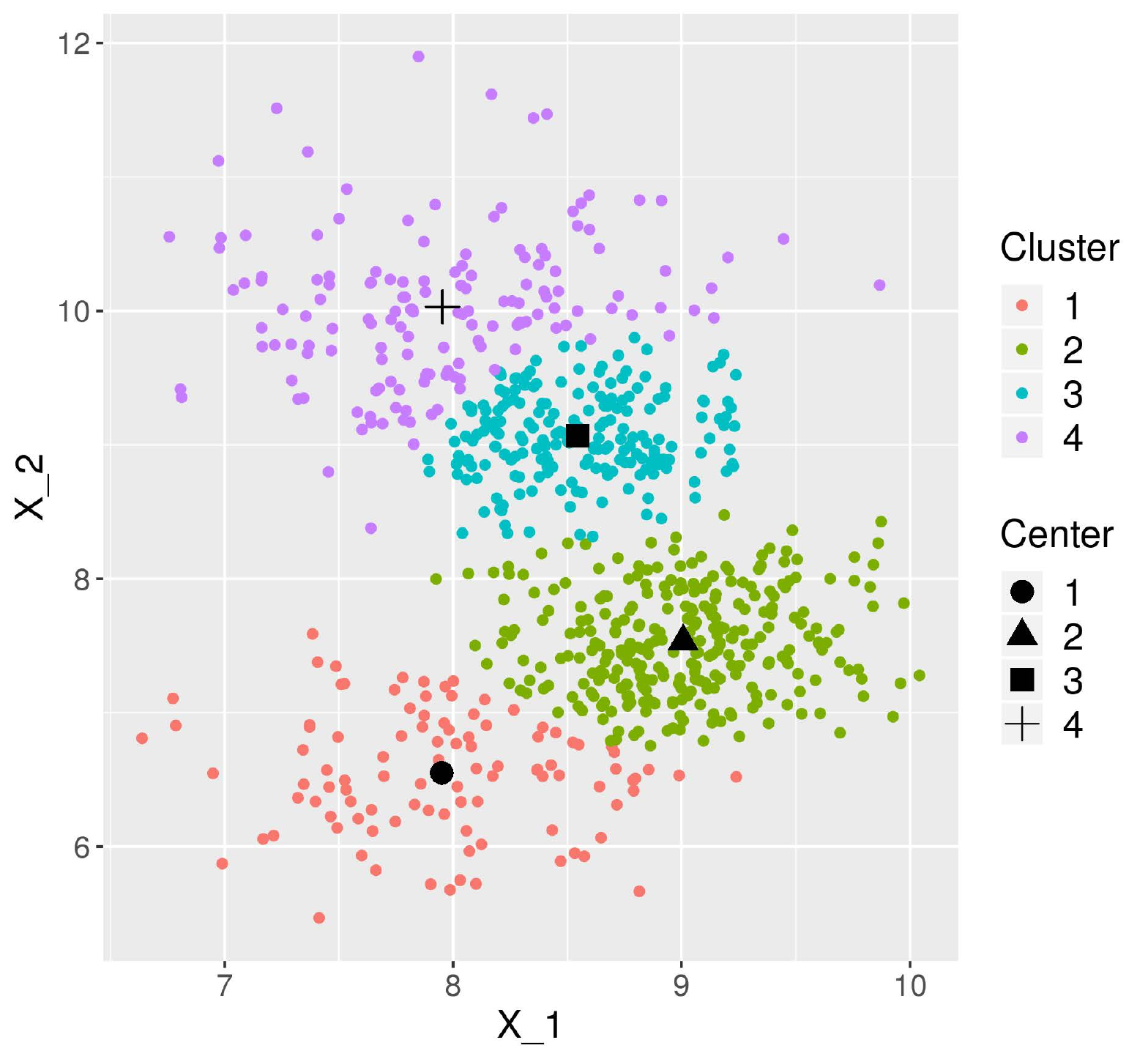}
                 }			
										
\caption{Clustering by the EM algorithm}   
\label{emclusters}   
\end{figure}

In the upper left panel of Figure \ref{emclusters} we display the original clusters (the same as above) for the sake of easy visual comparison, as well as the clusters obtained by applying the EM clustering methodology. 

Form Table \ref{emclusters} it becomes clear how the distance between points changes plus the type of clustering algorithm determine the centers of the clusters.

	\begin{table} 
	\centering\rowcolors{2}{gray!6}{white}
		\begin{tabular}{ccccc}	 \hline 
		\hiderowcolors
	Cluster & 1 & 2 & 3 & 4\\  \hline \hline
		\showrowcolors
		Original & (7.93, 6.63) & (9.01, 7.51) & (8.5, 9.08) & (8.01, 10.01)\\
		$h(x) = x$ & (7.82, 6.52) & (8.99, 7.49) & (8.47, 9.2) & (7.79, 10.29)\\
		$h(x) = \sqrt(x)$ & (7.72, 6.41) & (8.94, 7.44) & (8.26, 9.32) & (8.49, 10.53)\\
		$h(x) = \ln(x)$ & (7.95, 6.55) & (9.01, 7.53) & (8.55, 9.07) & (7.95, 10.03)\\
		$h(x) = \exp(0.5x)$ & (7.79, 6.59) & (8.88, 7.37) & (9.55, 7.67) & (8.31, 9.34)\\
		$h(x) = \exp(-0.5x)$ & (7.92, 6.6) & (9.01, 7.53) & (8.52, 9.1) & (7.93, 10.09)\\
	\end{tabular}
	\rowcolors{2}{white}{white}
		\caption{centers of EM clusters in the different metrics}
	\label{emcenter}
\end{table}

In Table (\ref{emsize}), we list the size of the original clusters along with the size of the clusters obtained in the different distances. Also, in the last column of the table we list the accuracy of the procedure,  defined as the number of cases correctly classified. Obviously, with the original data all the cases are correct, but note that since the distance between the points changes with the metric, it is natural that the accuracy of the procedure would vary.

\begin{table} 
	\centering\rowcolors{2}{gray!6}{white}	
	\begin{tabular}{cccccc}
		\hiderowcolors
		\hline
		Cluster & 1 & 2 & 3 & 4 & \textbf{Accuracy} \\ \hline \hline  
		\showrowcolors	
		Original     & 100 & 300 & 200 & 150 & 1\\
		$h(x) = x$           & 85  & 315 & 259 & 91  & 0.901\\
		$h(x) = \sqrt(x)$     & 64  & 333 & 307 & 46  & 0.809\\
		$h(x) = \ln(x)$       & 101 & 298 & 200 & 151 & 0.996\\
		$h(x) = \exp(0.5x)$   & 83  & 217 & 69 & 381  & 0.577\\
		$h(x) = \exp(-0.5x)$  & 101 & 303 & 208 & 138 & 0.977\\
	\end{tabular}
	\rowcolors{2}{white}{white}
    \caption{Size of the EM clusters and accuracy for the different metrics}
	\label{emsize}
\end{table}

\section{Clustering using the HCPC algorithm}
The hierarchical clustering on principal components (HCPC) algorithm is a combination of methods. First it uses principal component analysis (PCA) to reduce possible hidden high dimensionality in the data. Then it uses hierarchical and partitional clustering  to describe similarity between the points. See \cite{HJP} for details and further references.

\begin{figure} 
 \centering
	 \subfloat[Data set]{
                \includegraphics[width=2.0in,height=2.0in]{originalclusters.pdf}
               }
    \subfloat[$h(x)=x$]{
                \includegraphics[width=2.0in,height=2.0in]{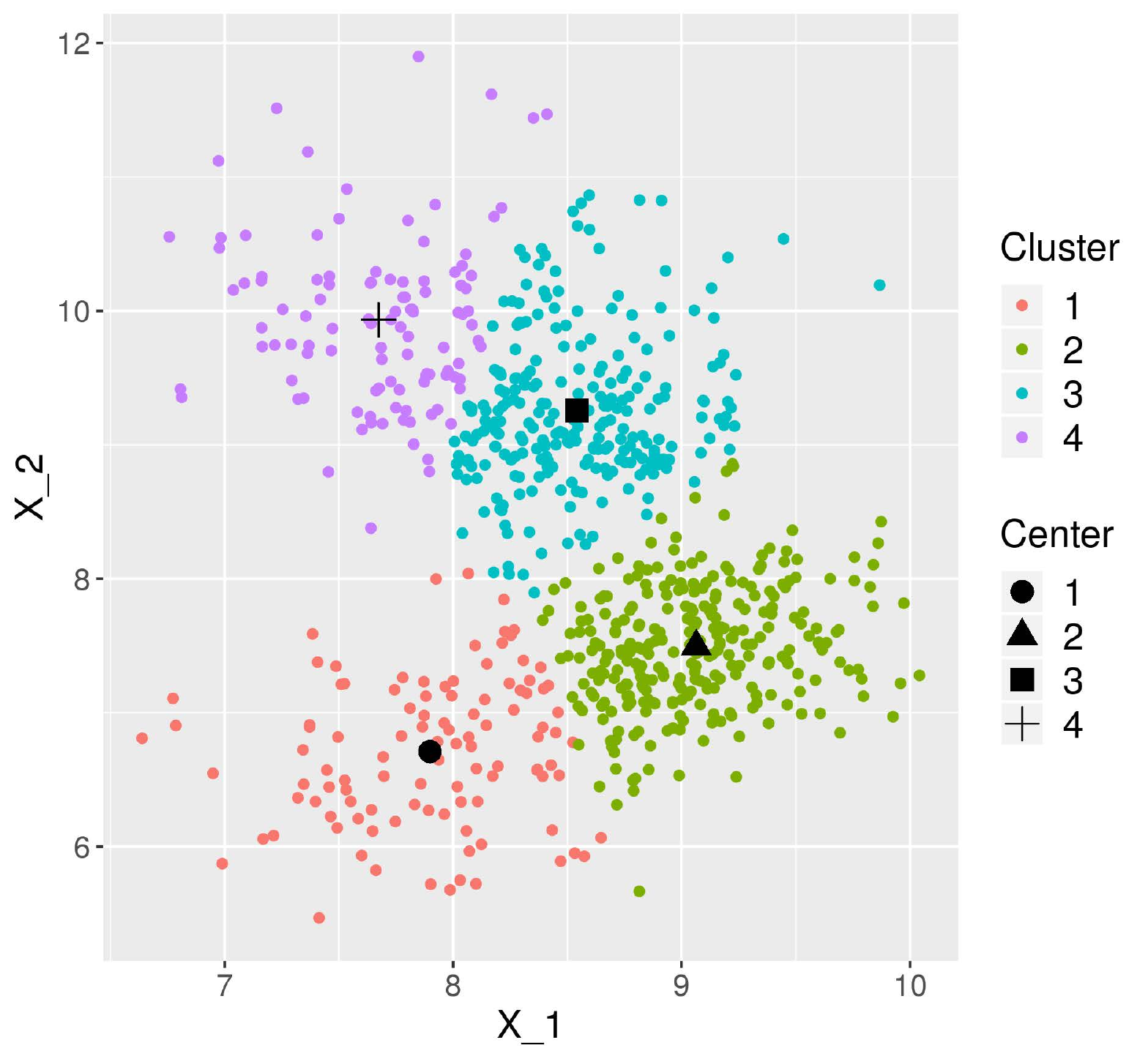}
                 }
	\subfloat[$h(x)=\exp(-0.5x)$]{
                \includegraphics[width=2.0in,height=2.0in]{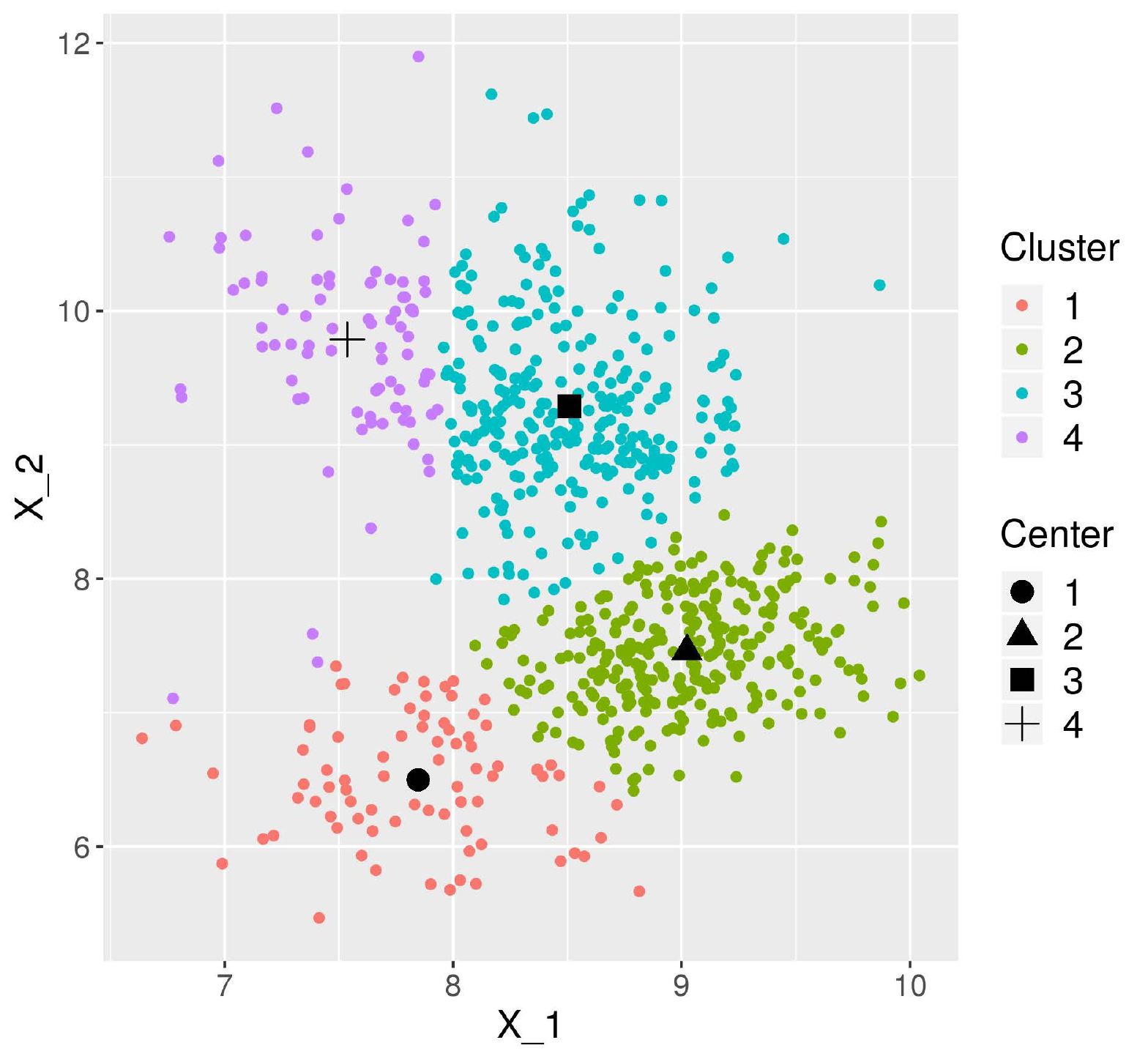}
                 }
								
  \subfloat[$h(x)=\exp(0.5x)$]{
                \includegraphics[width=2.0in,height=2.0in]{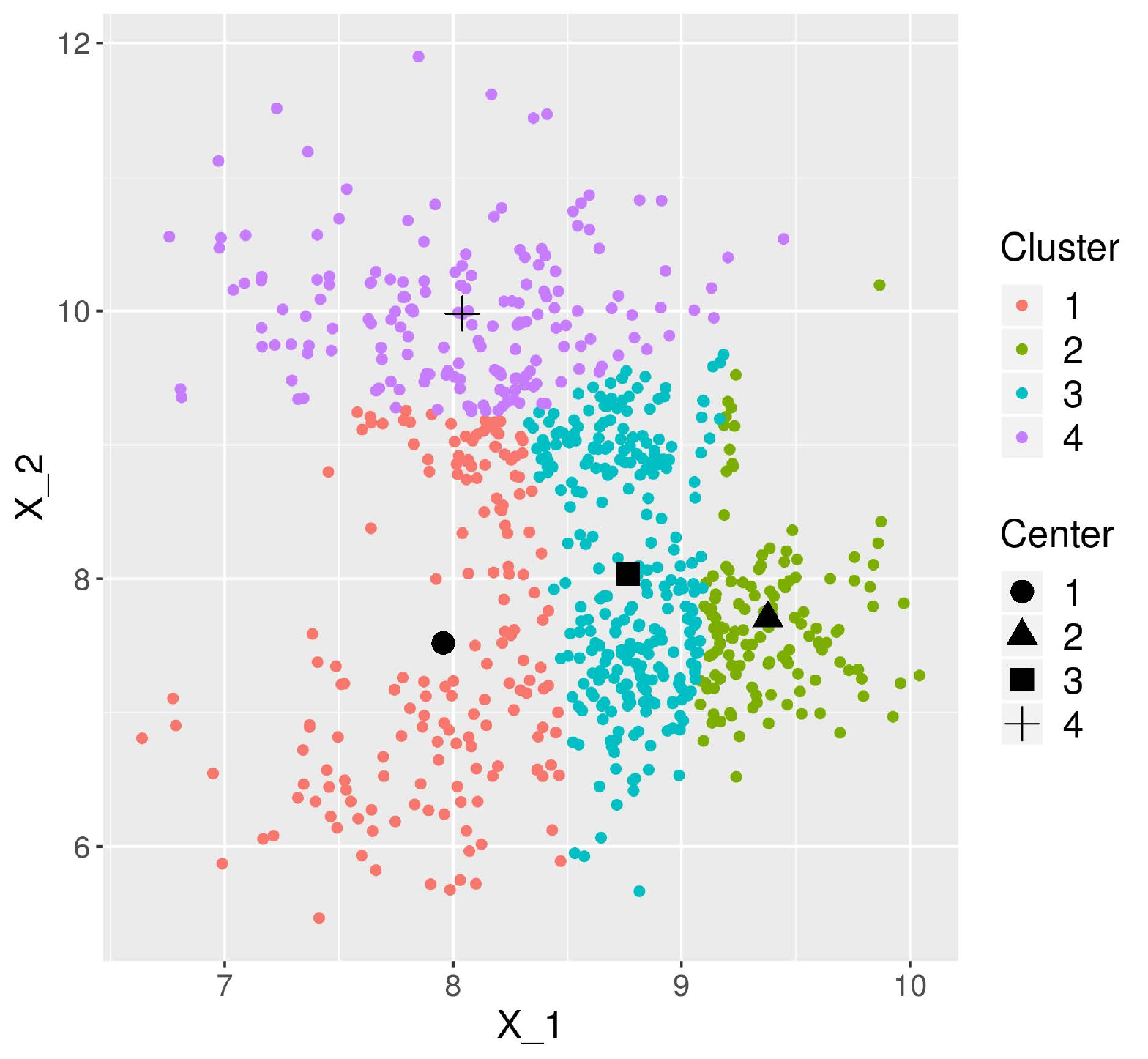}
               }
    \subfloat[$h(x)=\sqrt{x}$]{
                \includegraphics[width=2.0in,height=2.0in]{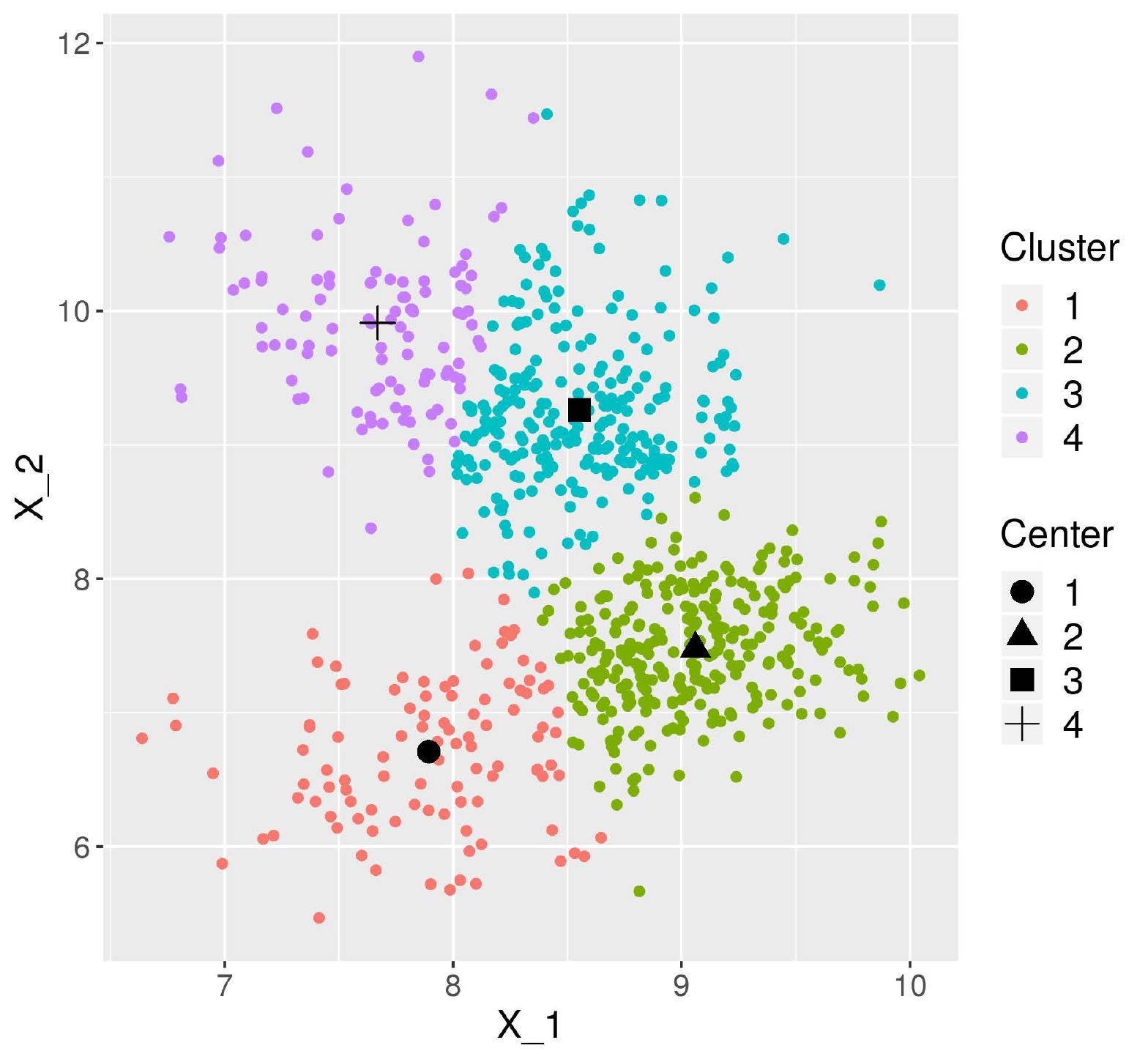}
                 }						
			   \subfloat[$h(x)=\ln x$]{
                \includegraphics[width=2.0in,height=2.0in]{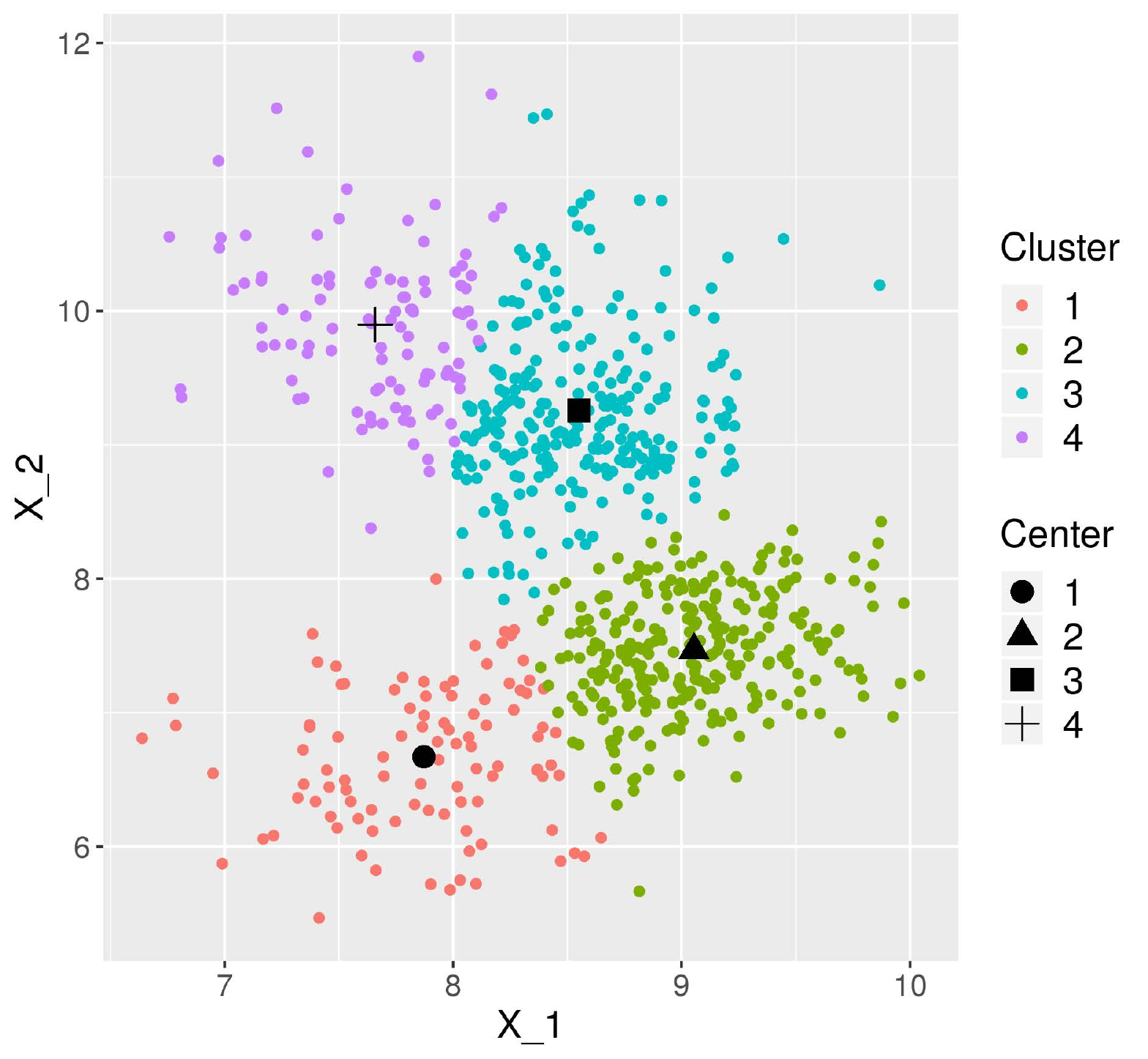}
                 }					
\caption{Clustering by the HCPC algorithm}   
\label{hcpcclusters}   
\end{figure}

In Figure \ref{hcpcclusters} we display the results of the clustering procedure. The panels are as displayed as above. In the upper leftmost one, appears the original data, and the remaining panels we display the results of the clustering process. The resulting clusters are in different colors.

In Table \ref{hcpccenters} we list the centers of the HCPC-clusters in the different distances.
	\begin{table} 
	\centering\rowcolors{2}{gray!6}{white}
	\begin{tabular}{ccccc}
		\hiderowcolors
		 \hline 
		Cluster & 1 & 2 & 3 & 4\\  \hline \hline
		\showrowcolors
		original & (7.93, 6.63) & (9.01, 7.51) & (8.5, 9.08) & (8.01, 10.01)\\
		$h(x) = x$ & (7.9, 6.71) & (9.06, 7.49) & (8.54, 9.26) & (7.67, 9.93)\\
		$h(x) = \sqrt(x)$ & (7.89, 6.71) & (9.06, 7.48) & (8.55, 9.26) & (7.67, 9.91)\\
		$h(x) = \ln(x)$ & (7.87, 6.67) & (9.05, 7.47) & (8.55, 9.25) & (7.66, 9.9)\\
		$h(x) = \exp(0.5x)$ & (7.96, 7.52) & (9.38, 7.71) & (8.77, 8.04) & (8.04, 9.98)\\
		$h(x) = \exp(-0.5x)$ & (7.85, 6.5) & (9.02, 7.46) & (8.51, 9.29) & (7.54, 9.79)\\
	\end{tabular}
	\rowcolors{2}{white}{white}
		\caption{centers of the HCPC clusters in the different metrics}
	\label{hcpccenters}
\end{table}

In Table \ref{hcpcsize}, we can observe the size of the original clusters along with the size obtained with the transformations. Also, this table contains a column that is the accuracy, this value is defined as the number of cases correctly classified. Obviously, with the original data all the cases are correct, for that reason we have a accuracy of 1, but for the other transformations we fail.

\begin{table} 
	\centering\rowcolors{2}{gray!6}{white}	
	\begin{tabular}{cccccc}
		\hiderowcolors
		\hline
		Cluster & 1 & 2 & 3 & 4 & \textbf{Accuracy} \\ \hline \hline  
		\showrowcolors
		original             & 100 & 300 & 200 & 150 & 1\\
		$h(x) = x$           & 106 & 290 & 247 & 107 & 0.929 \\
		$h(x) = \sqrt(x)$     & 105 & 288 & 250 & 107 & 0.927\\
		$h(x) = \ln(x)$       & 100 & 289 & 256 & 105 & 0.925\\
		$h(x) = \exp(0.5x)$   & 166 & 135 & 276 & 173 & 0.749\\
		$h(x) = \exp(-0.5x)$  & 83 & 297 & 285 & 85   & 0.864\\

	\end{tabular}
	\rowcolors{2}{white}{white}
    \caption{Size of the HCPC clusters and accuracy obtained with different transformations}
	\label{hcpcsize}
\end{table}

%%%%%%%%%%%
\section{The Hierarchical Agglomerative Clustering (HAC) process}
%%%%%%%%%%%%

The Hierarchical Agglomerative Clustering (HAC) is designed to build a hierarchy of clusters using a bottom-up or agglomerative strategy. Here each data point is treated as a single cluster and then a distance (or similarity) between each of the clusters is computed in order to merge two or more into one (e.g., single linkage or average linkage), that is, the two most ``similar'' clusters are merged at each step. See \cite{K} and \cite{N} for details.

\begin{figure} 
 \centering
	 \subfloat[Data set]{
                \includegraphics[width=2.0in,height=2.0in]{originalclusters.pdf}
               }
    \subfloat[$h(x)=x$]{
                \includegraphics[width=2.0in,height=2.0in]{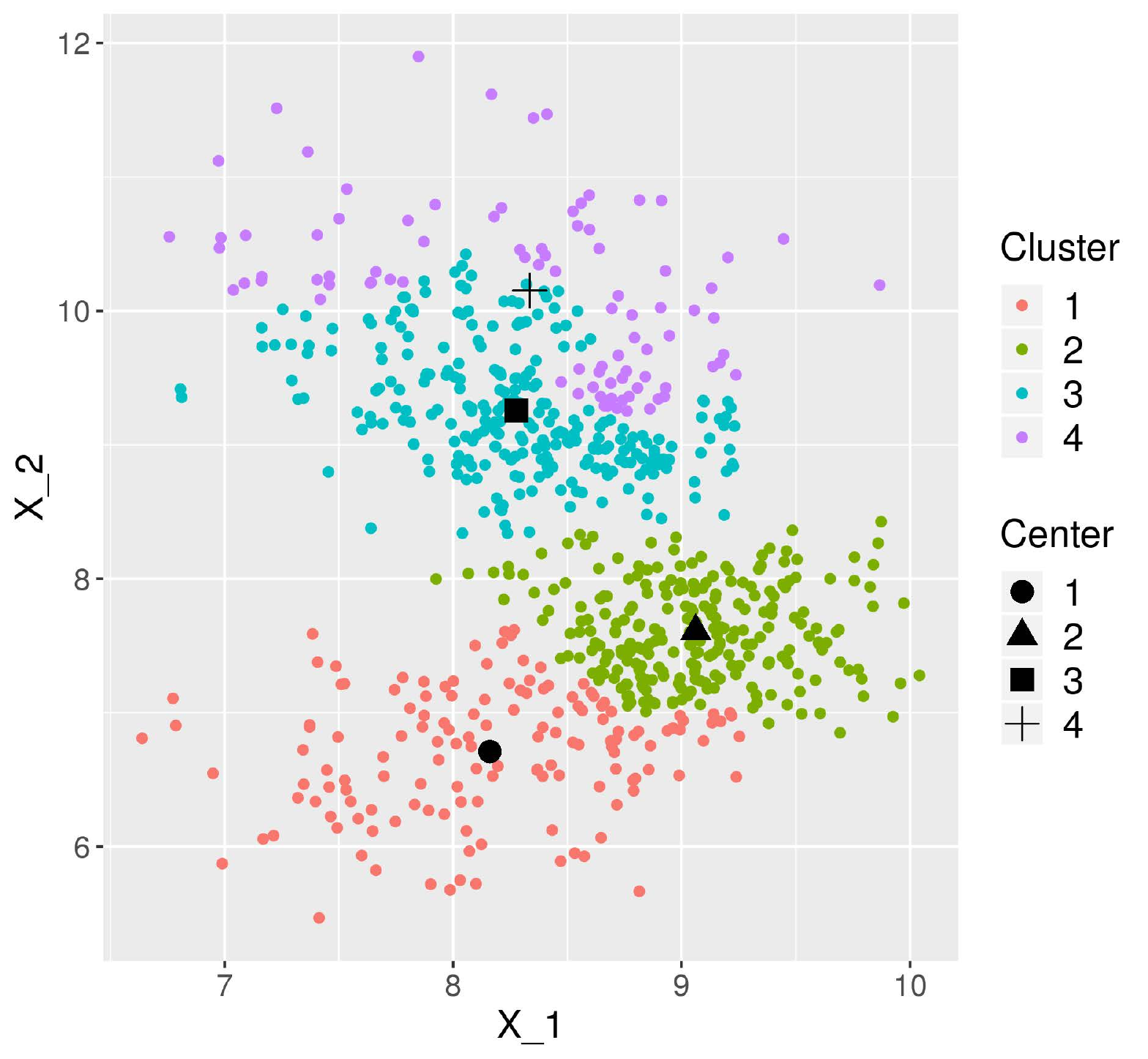}
                 }
	\subfloat[$h(x)=\exp(-0.5x)$]{
                \includegraphics[width=2.0in,height=2.0in]{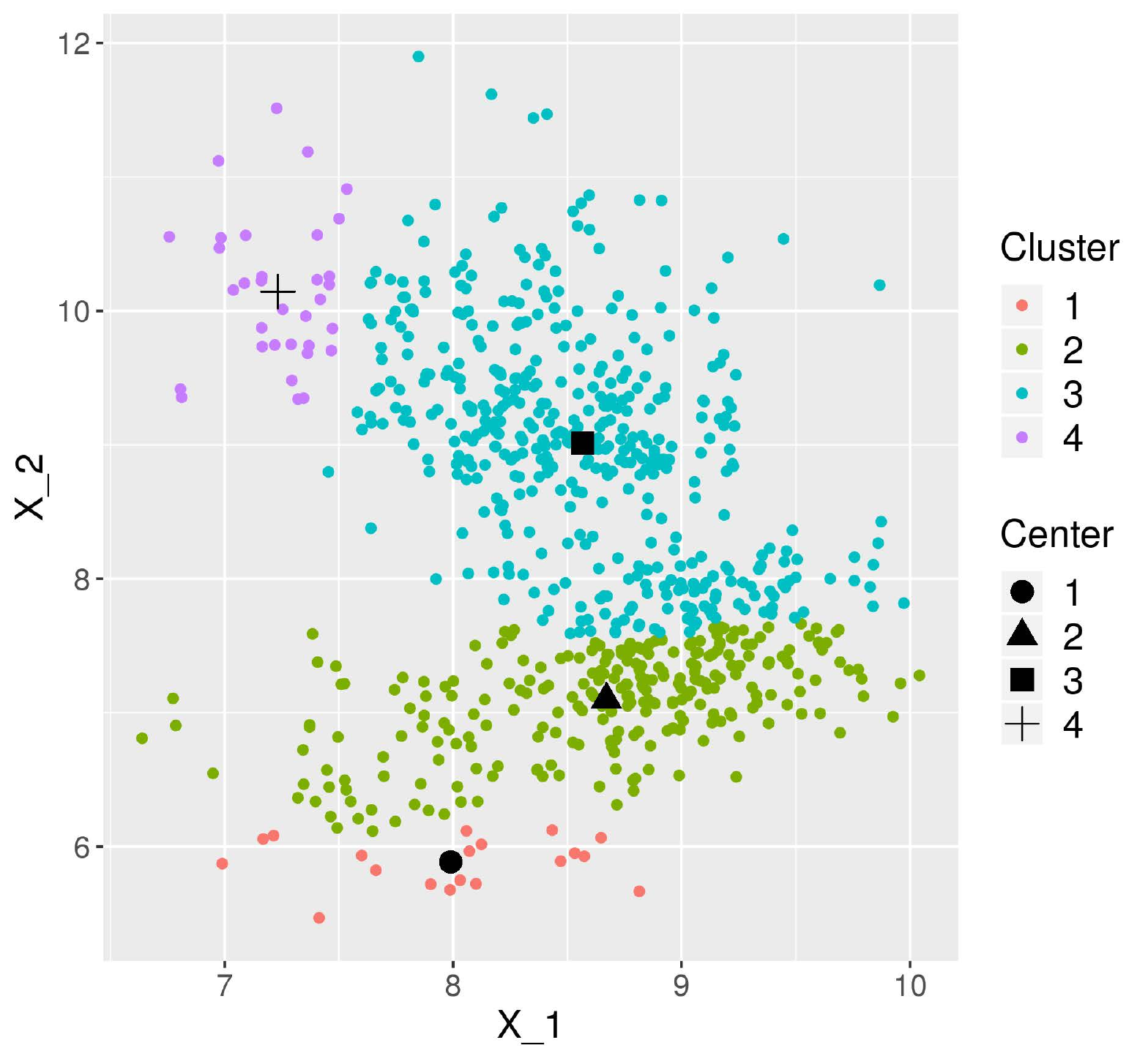}
                 }
								
  \subfloat[$h(x)=\exp(0.5x)$]{
                \includegraphics[width=2.0in,height=2.0in]{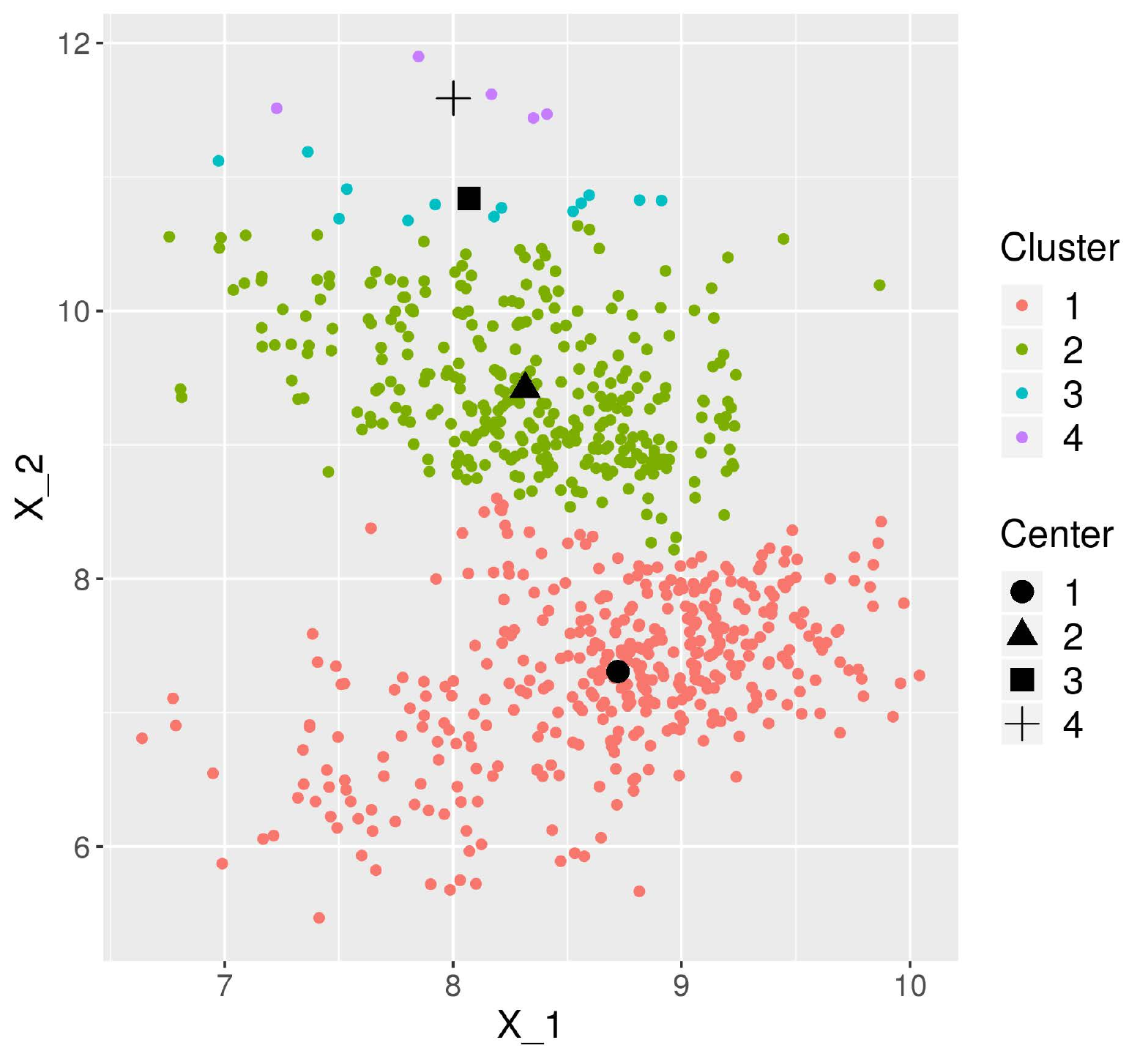}
               }
    \subfloat[$h(x)=\sqrt{x}$]{
                \includegraphics[width=2.0in,height=2.0in]{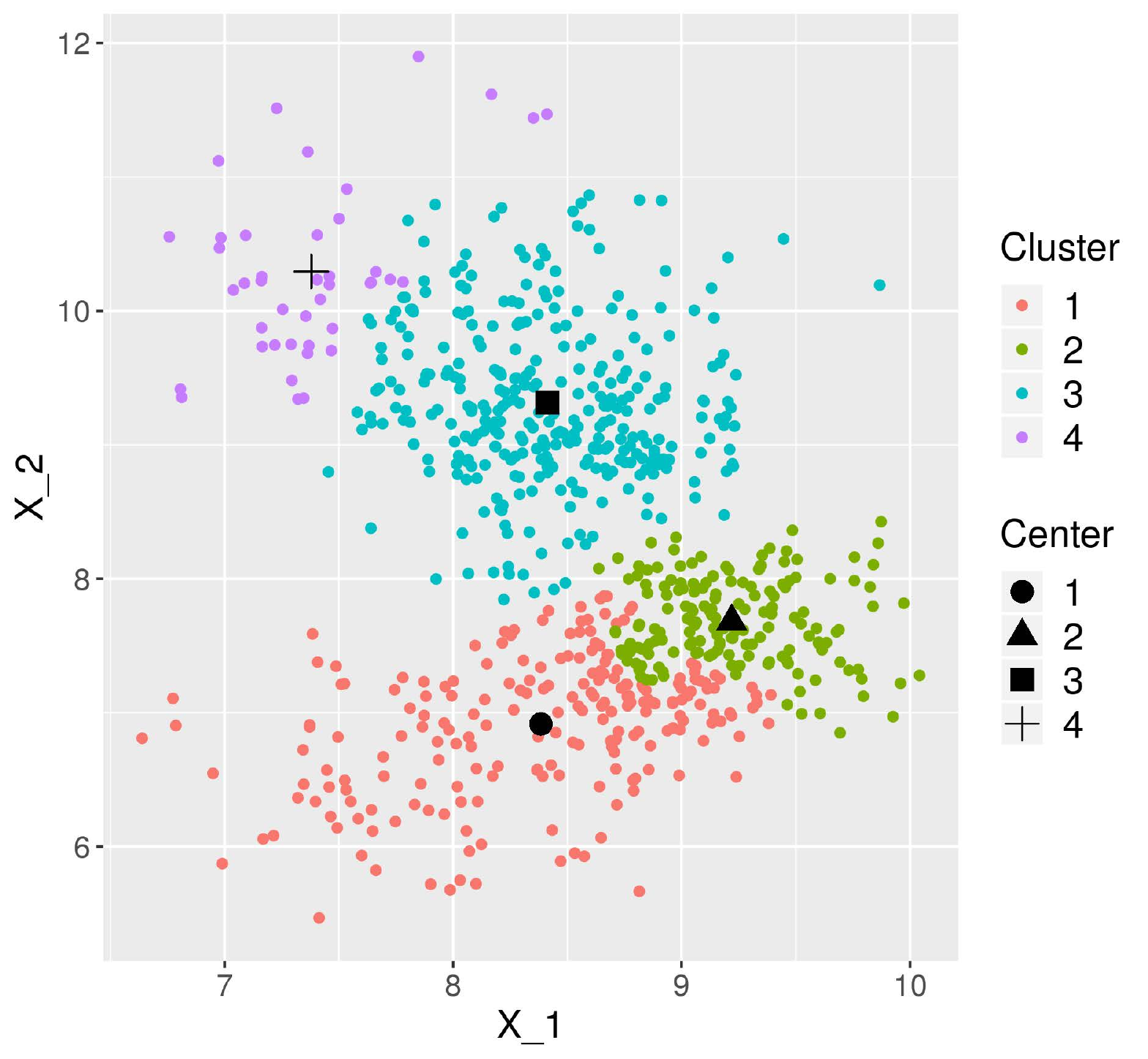}
                 }						
			   \subfloat[$h(x)=\ln x$]{
                \includegraphics[width=2.0in,height=2.0in]{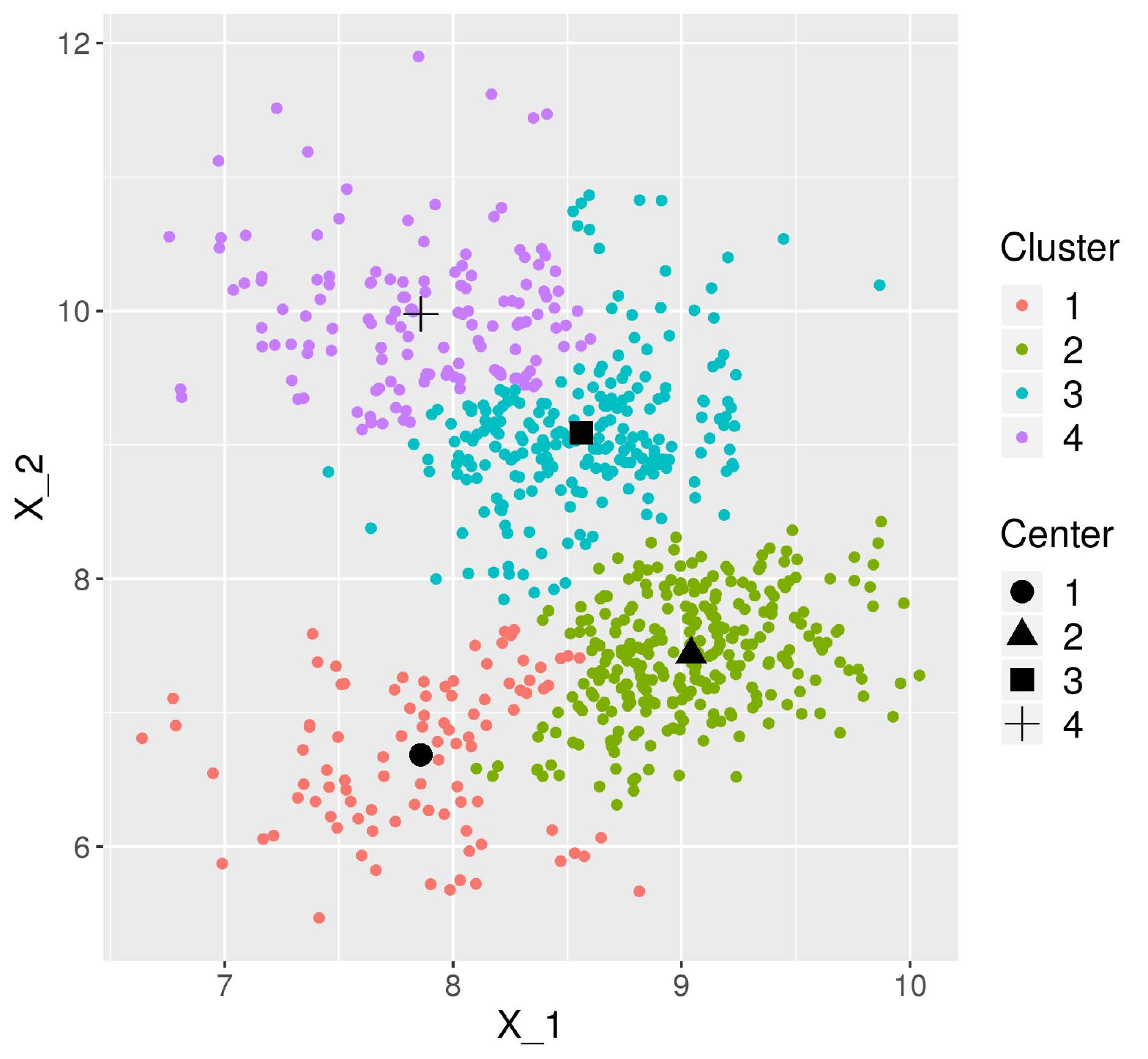}
                 }					
\caption{Clustering by the hierarchical clustering algorithm}  
\label{hclusters}   
\end{figure}

In Figure \ref{hclusters} we display the results of the hierarchical clustering algorithm.
The panels are organized as above.

In Table \ref{hcenter} we list the centers associated to the resulting clusters. As usual, they are the $d_\phi-$arithmetic means of the clusters provided by the hierarchical clustering algorithm.

\begin{table} 
	\centering\rowcolors{2}{gray!6}{white}
	\begin{tabular}{ccccc}
		\hiderowcolors
		 \hline 
		Cluster & 1 & 2 & 3 & 4\\  \hline \hline
		\showrowcolors
		original & (7.93, 6.63) & (9.01, 7.51) & (8.5, 9.08) & (8.01, 10.01)\\
		$h(x) = x$ & (8.16, 6.71) & (9.06, 7.61) & (8.28, 9.26) & (8.34, 10.15)\\
		$h(x) = \sqrt(x)$ & (8.38, 6.92) & (9.22, 7.68) & (8.41, 9.31) & (7.38, 10.29)\\
		$h(x) = \ln(x)$ & (7.86, 6.69) & (9.04, 7.44) & (8.56, 9.09) & (7.86, 9.98)\\
		$h(x) = \exp(0.5x)$ & (8.72, 7.31) & (8.32, 9.42) & (8.07, 10.84) & (8, 11.59)\\
		$h(x) = \exp(-0.5x)$ & (7.99, 5.89) & (8.67, 7.1) & (8.57, 9.01) & (7.23, 10.14)\\
	\end{tabular}
	\rowcolors{2}{white}{white}
		\caption{centers of the clusters provided by the hierarchical clustering }
	\label{hcenter}
\end{table}

In Table \ref{hsize}, we list  the size of the original clusters along with the sizes of the clusters in the different metrics. We also list the accuracy of the classifier. 

{\bf Comment} At this point we mention that in all examples the accuracy can be provided because we new the true clusters before hand. Therefore, this accuracy measure is an indirect quality test of the different clustering methods.
Other measures of cluster quality can be considered (e.g., the adjusted Rand index or the Normalized Mutual Information~\cite{BBC2}).

\begin{table} 
	\centering\rowcolors{2}{gray!6}{white}	
	\begin{tabular}{cccccc}
		\hiderowcolors
		\hline
		Cluster & 1 & 2 & 3 & 4 & \textbf{Accuracy} \\ \hline \hline  
		\showrowcolors
		original             & 100 & 300 & 200 & 150 & 1\\
		$h(x) = x$           & 144 & 256 & 261 & 89  & 0.860\\
		$h(x) = \sqrt(x)$     & 215 & 170 & 323 & 42  & 0.683\\
		$h(x) = \ln(x)$       & 94 & 291 & 226 & 139  & 0.957\\
		$h(x) = \exp(0.5x)$   & 407 & 325 & 13 & 5    & 0.140\\
		$h(x) = \exp(-0.5x)$  & 19 & 263 & 435 & 33   & 0.579\\
	\end{tabular}
	\rowcolors{2}{white}{white}
    \caption{Size of the Hierarchical clustering and accuracy obtained with different transformations}
	\label{hsize}
\end{table}

%%%%%%%
\section{Concluding comments}
%%%%%%%
In this note, we considered the Riemannian metric distance induced by a separable Bregman divergence, and studied the fixed rate quantization problem. We concisely described the Riemann-Bregman Voronoi diagrams that can be obtained from an ordinary Euclidean Voronoi diagram after a monotone isometric embedding. 
Then we reported on the partition-based ($k$-means), soft (EM) and hierarchical (HCPC/HCA) clusterings with respect to these Riemann-Bregman distances. As was to be expected, the resulting clusters depend on the clustering method as well as on the metric used to measure the distance between points.

\end{document}